\documentclass[letterpaper]{article} 
\usepackage{aaai25}  
\usepackage{times}  
\usepackage{helvet}  
\usepackage{courier}  
\usepackage[hyphens]{url}  
\usepackage{graphicx} 
\usepackage{natbib}  
\usepackage{caption} 
\usepackage{algorithm}
\usepackage{algorithmic}

\usepackage{newfloat}
\usepackage{listings}
\DeclareCaptionStyle{ruled}{labelfont=normalfont,labelsep=colon,strut=off} 
\floatstyle{ruled}
\newfloat{listing}{tb}{lst}{}
\floatname{listing}{Listing}
\usepackage{multirow}
\usepackage{xcolor}
\usepackage{amsfonts}
\usepackage{amsmath}
\usepackage{subcaption}
\usepackage{cleveref}
\usepackage{labelcas}

\DeclareMathOperator*{\argmin}{arg\,min}
\DeclareMathOperator*{\argmax}{arg\,max}

\title{How not to Stitch Representations to Measure Similarity:\\ Task Loss Matching versus Direct Matching}
\author{
    András Balogh\textsuperscript{\rm 1},
    Márk Jelasity\textsuperscript{\rm 1,2}
}
\affiliations{
    \textsuperscript{\rm 1} University of Szeged, Hungary\\
    \textsuperscript{\rm 2} HUN-REN-SZTE Research Group on AI, Szeged, Hungary\\
    \{abalogh, jelasity\}@inf.u-szeged.hu
}

\begin{document}

\maketitle

\begin{abstract}
Measuring the similarity of the internal representations of deep neural networks
is an important and challenging problem.
Model stitching has been proposed as a possible approach, where two half-networks
are connected by mapping the output of the first half-network to the input of the
second one.
The representations are considered functionally similar if the resulting stitched
network achieves good task-specific performance.
The mapping is normally created by training an affine stitching layer
on the task at hand while freezing the two half-networks, a method called task
loss matching.
Here, we argue that task loss matching may be very misleading as a similarity index.
For example, it can indicate very high similarity between very distant
layers, whose representations are known to have different functional properties.
Moreover, it can indicate very distant layers to be more similar than architecturally corresponding
layers.
Even more surprisingly, when comparing layers within the same network, task loss
matching often indicates that some layers are more similar to a layer than itself.
We argue that the main reason behind these problems is that task loss matching
tends to create out-of-distribution representations to improve task-specific performance.
We demonstrate that direct matching (when the mapping minimizes the
distance between the stitched representations) does not suffer from these problems.
We compare task loss matching, direct matching, and well-known similarity
indices such as CCA and CKA. We conclude that direct matching strikes a good
balance between the structural and functional requirements for a good similarity
index.
\end{abstract}

%
\begin{links}
    \link{Code}{https://github.com/szegedai/stitching-ood}
\end{links}

\section{Introduction}

Understanding the internal representations that emerge as a result of training machine learning models
is a crucial problem, and an important aspect of this problem is to measure the similarity
of representations~\cite{kornblith2019similarity}.

However, this is a poorly defined problem, because similarity can be approached from many angles.
One choice is \emph{structural similarity} based on the geometric properties of the distribution of the
activations of, for example, a set of neurons of interest~\cite{kornblith2019similarity,raghu2017svcca,morcos2018insights}.
Another approach is \emph{functional similarity}, that is,
to ask the question whether one representation can be transformed into the other while preserving its
functionally important aspects~\cite{csiszarik2021similarity,bansal2021revisiting}.

Functional similarity has the great advantage that it grounds the meaning of similarity.
A purely structural measure conveys very little information on whether the two compared representations
are actually compatible or interchangeable in any practical sense~\cite{ding2021grounding}.
However, we argue that purely functional similarity approaches are not a panacea.

\subsection{Contributions}

\paragraph{Drawbacks of functional similarity.}
We show that a purely functional notion of similarity also has significant
drawbacks.
Not taking the structural properties into
account results in anomalies such as representations of one model being transformed into
highly out-of-distribution representations in the other model that do perform well functionally but
strongly violate some basic sanity checks required from any similarity index, such as identifying architecturally
corresponding layers.

\paragraph{A hybrid similarity index.}
We argue that for measuring similarity, the most promising approach is to combine structural and functional
aspects.
We examine one such approach: model stitching based on the direct matching of representations.
We find that it is a good compromise between purely structural and
purely functional approaches.

\subsection{Related work}

\subsubsection{Structural Similarity.}
Several methods have been proposed to quantitatively compare the learned internal representations of neural networks
based on geometrical and statistical properties of the distribution of activations.
For example, centered kernel alignment \cite{kornblith2019similarity}, orthogonal Procrustes distance \cite{schonemann1966generalized} and methods based on canonical correlation analysis \cite{raghu2017svcca, morcos2018insights} have been extensively used to analyze and compare representations \cite{smith2017offline, raghu2021do, yadav2024masked}.
However, these structural similarity indices do not take the functionality of the networks into account directly.
Accordingly, \cite{ding2021grounding} and \cite{hayne2024does} both show that these indices often fail in functional benchmarks.
For example, they might not be sensitive to changes that strongly affect predictive performance, and they might be overly sensitive to non-functional differences.

\subsubsection{Functional Similarity.}
Model stitching was introduced by \cite{lenc2015understanding}, who used $1 \times 1$ convolutional stitching layers to connect two ``half-networks'' in order to study the equivalence of their learned representations.
\cite{csiszarik2021similarity} proposed using model stitching as a framework to study the similarity of representations from a functional point of view.
Along with \cite{bansal2021revisiting}, they show that networks that are trained on the same task but under different settings can be stitched with minimal performance loss, which suggests that all successful networks achieve functionally similar representations.
Model stitching has also been used to functionally compare representations between architecturally different models \cite{mcneely2020inception, yang2022deep} and between robust and non-robust networks \cite{jones2022if, balogh2023functional}.

Recently, the reliability of stitching for analyzing similarity has been questioned by \cite{hernandez2022model}, who showed that, in many cases, distant layers of classifiers can be stitched with high accuracy.
Our work sheds light on the reason behind anomalies of this type.

\subsubsection{Applied stitching.}
Many works have applied stitching as a method of combining networks, not focusing on similarity \cite{teerapittayanon2023stitchnet, guijt2024stitching}.
\cite{pan2023stitchable, he2024efficient, pan2023stitched} combine models of different sizes with stitching to achieve the optimal performance at various resource constraints.
Our work suggests that the success of such applications is not necessarily due to some underlying similar representations.

\section{Preliminaries}

Here, we summarize the basic concepts related to model stitching and also introduce the similarity indices that we will use in our comparisons later on.

\subsection{Functional Similarity: Model Stitching}

Our notation closely follows that of \cite{balogh2023functional}.
Let $f: \mathcal{X} \rightarrow \mathcal{Y}$ be a feedforward neural network with $m$ layers: $f = f_m \circ \cdots \circ f_1$ where $f_i: \mathcal{A}_{f, i-1} \rightarrow \mathcal{A}_{f, i}$ maps the activation space of layer $i-1$ to that of layer $i$. By definition, $\mathcal{A}_{f, 0} = \mathcal{X}$.
Since model stitching is a framework where two ``half-networks'' are connected, we introduce the notations $f_{\leq i} = f_i \circ \cdots \circ f_1$ and $f_{>i} = f_m \circ \cdots \circ f_{i+1}$.

Given two frozen networks $f$ and $g$, and one layer from each network $f_i$ and $g_j$, the abstract goal of stitching is to find out if $g_{>j}$, which we will refer to as the \textit{receiver}, can achieve its function using the representation of $f_{\leq i}$, which we will call the \textit{source}. In examining this, we attempt to find a transformation $T: \mathcal{A}_{f, i} \rightarrow \mathcal{A}_{g, j}$ such that the composite (or stitched) network $g_{>j} \circ T \circ f_{\leq i}$ is functionally similar to $g$.

Throughout this paper we will refer to $T$ as the \textit{stitching layer}, and for an input $x$, we will call $f_{\leq i}(x)$ its \textit{source representation} and $g_{\leq j}(x)$ its \textit{target representation}. Using these terms, the goal of stitching is to find a transformation that matches the source representation to the target. In the following, we will discuss the constraints on $T$ and our methods for finding the optimal transformation.

\subsubsection{The complexity of $T$.}
A sensible requirement on $T$ is to not increase the complexity of the stitched network significantly, but be expressive enough to allow for non-trivial mappings. The general consensus is that affine transformations are suitable \cite{csiszarik2021similarity, bansal2021revisiting} and we adopt this approach in this work.

For convolutional representations of sizes $\mathbb{R}^{c_1\times h\times w}$ and $\mathbb{R}^{c_2\times h\times w}$, where $c_1$ and $c_2$ are the number of feature maps, and $h$ and $w$ are the spatial dimensions of these feature maps, the stitching transformation is a $1 \times 1$ convolutional layer that applies the same affine mapping $M: \mathbb{R}^{c_1} \rightarrow \mathbb{R}^{c_2}$ to each of the $h \times w$ spatial positions.

If the representations have different spatial dimensions, a 2D resizing operation of each feature map precedes the stitching transformation.
We used bilinear interpolation.

For transformer embeddings of sizes $\mathbb{R}^{n\times d_1}$ and $\mathbb{R}^{n\times d_2}$, where $n$ is the number of tokens and $d_1$ and $d_2$ are the embedding dimensions,
we apply the same affine mapping $M: \mathbb{R}^{d_1} \rightarrow \mathbb{R}^{d_2}$ to every token.

\subsubsection{Quantifying functional similarity.}
Following \cite{csiszarik2021similarity}, the functional similarity of the two selected layers will be quantified by the stitched network's performance,
relative to the performance of the target network $g$. In our case, the functionality of interest is accuracy over a supervised classification problem,
given by an underlying distribution $p(x, y)$ over $\mathcal{X} \times \mathcal{Y}$.
Thus, similarity is characterized by relative accuracy.

\subsubsection{Task loss matching (TLM).}
Given that we treat functional performance as a measure of similarity, a natural way to find the optimal stitching transformation is through solving the learning task
\begin{equation}
    \argmin_{\theta} \mathbb{E}_{p(x,y)} [ \mathcal{L}([g_{>j} \circ T_\theta \circ f_{\leq i}](x), y) ]
    \label{eq:stitching-task}
\end{equation}
for a suitable surrogate loss function $\mathcal{L}: \mathcal{Y} \times \mathcal{Y} \rightarrow \mathbb{R}$, over a dataset $D = \{ (x_i, y_i) \}_{i=1}^n$ from the distribution $p(x, y)$. We only train the parameters $\theta$ of $T_{\theta}$, while $g_{>j}$ and $f_{\leq i}$ are frozen, as mentioned before.

\subsubsection{Direct matching (DM).}
\label{sec:direct_matching}
Another option is to directly match the source representation to the target through solving the optimization problem
\begin{equation}
    \argmin_{\theta} \mathbb{E}_{p(x)}[ \lVert [T_\theta \circ f_{\leq i}](x) -  g_{\leq j}(x) \rVert_F ].
    \label{eq:matching-task}
\end{equation}
Note that this problem does not depend on $y$, hence it is independent of the classification task.
Given that our choice of $T_\theta$ is a common affine transformation for each spatial position or token,
sampling a set of equations
\begin{equation}
[T_\theta \circ f_{\leq i}](x_k) = g_{\leq j}(x_k),\ \ x_k\sim p(x),\ k=1,\ldots,K,
\end{equation}
results in a linear problem in $\theta$.
We compute $\theta$ using the Moore-Penrose pseudoinverse.

\subsection{Structural Similarity Indices}

Structural similarity indices compare the internal activations of two layers over the same set of $n$ examples.
The goal is to quantify the similarity between two activation matrices $A \in \mathbb{R}^{n \times p_1}$ and $B \in \mathbb{R}^{n \times p_2}$, where $p_1$ and $p_2$ are the number of neurons.
Let us now very briefly introduce the basic notions of three popular similarity indices that we will use in our evaluation.
In the following, we assume that both $A$ and $B$ have centered columns.

\subsubsection{Linear CKA.} Centered kernel alignment (CKA)~\cite{cka}
measures the alignment of the representations' inter-example similarity structures, essentially computing an inner product of the
vectors of pairwise similarities among the $n$ examples in the two representations.
\cite{kornblith2019similarity} report similar results between the linear and RBF kernels used for measuring similarity between examples,
so in this paper we focus on the linear variant of CKA, which can be computed as
\begin{equation}
    \text{LCKA}(A, B) = \frac{\lVert B^T A\rVert_F^2}{\lVert A^T A\rVert_F\lVert B^T B\rVert_F}.
\end{equation}

\subsubsection{PWCCA.} Canonical Correlation Analysis (CCA) has also been used to compute similarity indices \cite{raghu2017svcca, morcos2018insights}.
The canonical correlation coefficients $\rho_i$ are given by
\begin{equation}
    \rho_i = \max_{\mathbf{w}_A^i, \mathbf{w}_B^i} \text{corr}(A\mathbf{w}_A^i, B\mathbf{w}_B^i)
\end{equation}
subject to the orthogonality conditions $\langle A\mathbf{w}_A^i, B\mathbf{w}_A^j\rangle=0$ and $\langle A\mathbf{w}_B^i, B\mathbf{w}_B^j\rangle=0$
for $j<i$.

The coefficients $\rho_i$ can be used to compute a similarity score.
The projection-weighted average of the coefficients was proposed by \cite{morcos2018insights}, which was shown to be the best performing
CCA variant in the benchmarks of \cite{ding2021grounding}, given by
\begin{equation}
    \text{PWCCA}(A, B) = \frac{\sum_i \alpha_i \rho_i}{\sum_i \alpha_i}, \; \; \; \alpha_i = \sum_j |\langle A \mathbf{w}_A^i,  \mathbf{a}_j \rangle |
\end{equation}
where $\mathbf{a}_j$ is the $j^\text{th}$ column of $A$.

\subsubsection{OPD.} The orthogonal Procrustes problem asks for an orthogonal transformation of $A$ such that it is closest to $B$ in the Frobenius norm:
\begin{equation}
    R^* = \argmin_R \lVert B - AR \rVert_F^2 \;\;\text{subject to} \;\; R^T R = I.
	\label{eq:rstar}
\end{equation}

The minimum distance (that we get when using the optimal transformation $R^*$), also known as the Orthogonal Procrustes Distance (OPD), can
be computed indirectly without computing $R^*$ as
\begin{equation}
    \text{OPD}(A, B) = \lVert A \rVert_F^2 + \lVert B \rVert_F^2 - 2\lVert B^T A \rVert_*
\end{equation}
where $\lVert \cdot \rVert_*$ denotes the nuclear norm. \cite{ding2021grounding} show that OPD consistently outperforms PWCCA and LCKA in their benchmarks.
Note that OPD is a distance function, not a similarity score, and we use it accordingly.

\subsection{Two-Faced Similarity Indices}
\label{sec:twoface}

Some of the methods can be used both as a functional and a structural similarity index.
Direct matching, for example, creates an approximation of the target representation using
the source representation, where the Frobenius norm of the approximation error can serve as a structural distance metric.
Similarly, the optimal orthogonal transformation $R^*$ in \cref{eq:rstar} can also be interpreted as a variant of direct matching,
using $AR^*$ as the input to the second half-network.
\cite{csiszarik2021similarity} have argued that this functional variant of OPD is not a promising approach.
However, structural direct matching is an interesting baseline that we will include in our evaluations.

\section{The Unreliability of Task Loss Matching}
\label{sec:sanity}

\begin{figure*}[t]
    \centering
    \begin{subfigure}[b]{0.28\textwidth}
        \centering
        \includegraphics[width=\textwidth]{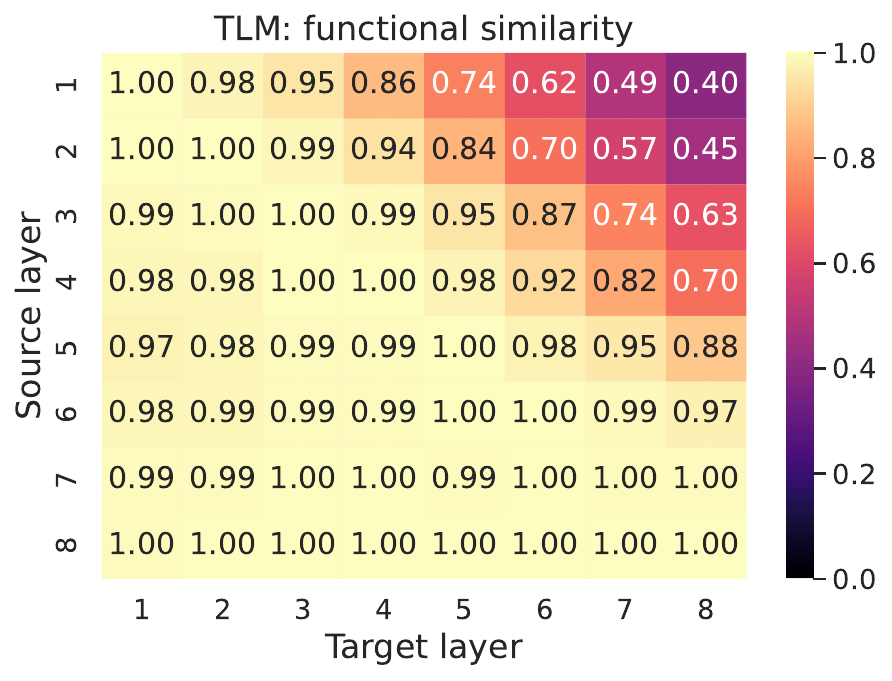}
        \caption{}
        \label{fig:self_stitch_sim_large}
    \end{subfigure}
    \begin{subfigure}[b]{0.28\textwidth}
        \centering
        \includegraphics[width=\textwidth]{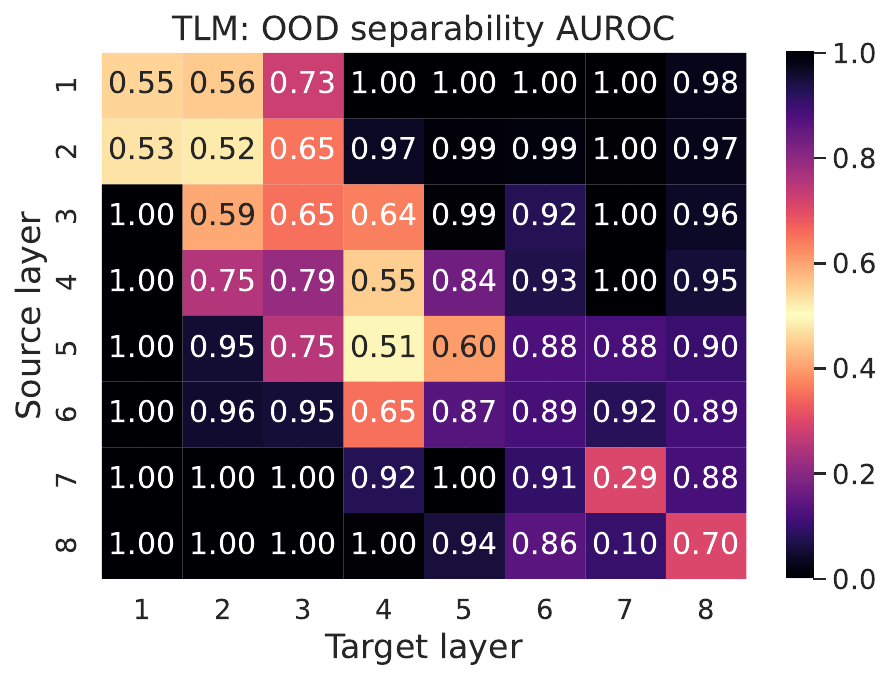}
        \caption{}
        \label{fig:self_stitch_ood_large}
    \end{subfigure}
    \begin{subfigure}[b]{0.43\textwidth}
        \centering
        \includegraphics[width=\textwidth]{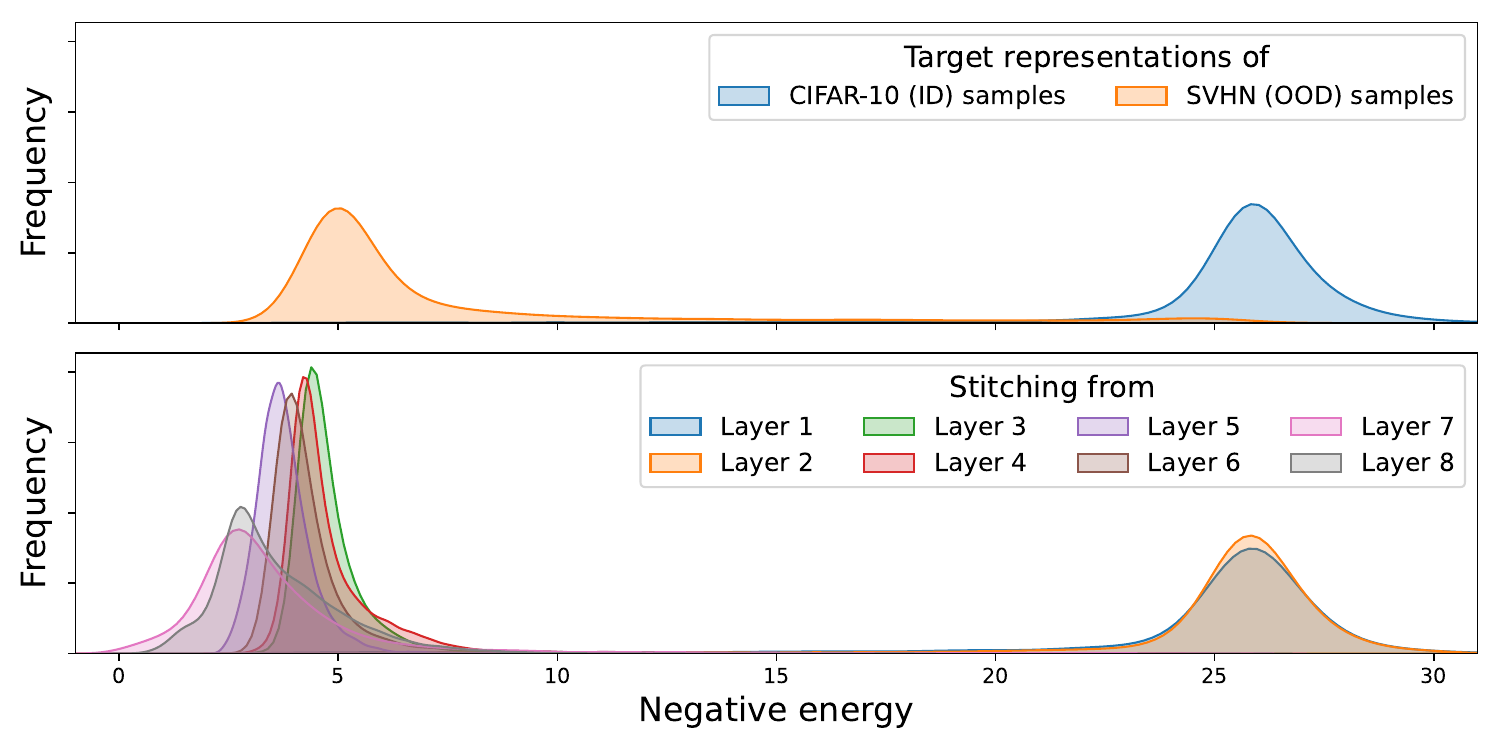}
        \caption{}
        \label{fig:self_stitch_energy_scores_large}
    \end{subfigure}
    \caption{ResNet-18, CIFAR-10, intra-network results with task loss matching.
	         (a) Pairwise similarities of layers (relative accuracy).
             (b) OOD level of the target representations (AUROC).
             (c) Energy distributions used by the OOD detector, when layer 1 is the target layer.
                 (c) top: energy distributions of layer 1 on in-distribution and OOD samples;
                 (c) bottom: energy distributions created by stitching from every layer, \emph{when measured
				 over in-distribution samples}. These distributions correspond to column 1 of the OOD
				 matrix in (b).}
    \label{fig:self_stitch_large}
\end{figure*}

Here, we will demonstrate that task loss matching fails the following two important sanity checks
rather spectacularly:
\begin{description}
\item[Inter-network layer identification.]
\cite{kornblith2019similarity} argue that
between two architecturally identical networks trained from different initializations,
for every layer in one network, \emph{the architecturally corresponding layer should
      be the most similar one} in the other network.
They showed that, by far, LCKA performs best in this test.
\item[Intra-network layer identification.]
We verify the seemingly obvious requirement that within one network,
for every layer the most similar layer should be itself.
\end{description}

\paragraph{Self-stitching.}
Previous studies, such as \cite{Csiszarik2021a, Bansal2021a, balogh2023functional} focused on model stitching between the corresponding layers of architecturally identical networks.
However, the concept of stitching is not restricted to corresponding layers or different networks.
Here, to test the intra-network sanity check, we will use \textit{self-stitching} where the stitching layer connects layers from the same network.
This essentially means cutting out layers, or replicating layers within a network if the two stitched layers are not the same.
If they are the same, self-stitching simply inserts a stitching transformation after the given layer.

\subsection{Experimental setup.}
We conduct experiments with two types of image classifiers: ResNets \cite{deepresidual16} and Vision Transformers (ViTs) \cite{Dosovitskiy2021a}.
We used the ResNet-18 and ViT-Ti architectures in our experiments.
We trained 10 networks of each type on the CIFAR-10 \cite{Krizhevsky2009a} dataset with identical hyperparameters, but different random initializations.

The layers we considered were all the residual blocks and the transformer encoder blocks of our ResNets and ViTs, respectively.
All stitchers were initialized using the direct matching method described in \cref{sec:direct_matching} with $K=100$ samples from the CIFAR-10 training set.
Task loss matching was then trained for 30 epochs with identical parameters as in \cite{Csiszarik2021a}.
All the similarity indices were evaluated over the CIFAR-10 test set.
For further training details, please see the Supplementary.

\subsection{Results.}
We measured intra-network similarities for all networks, and inter-network similarities between all pairs of networks.
\Cref{tab:matching_layers} contains the accuracies of identifying the architecturally corresponding layers based on maximum similarity.

\begin{table}[t]
\centering
\begin{tabular}{lcccc}
\hline
\multicolumn{1}{c}{\multirow{2}{*}{\begin{tabular}[c]{@{}c@{}}Similarity \\ index\end{tabular}}} & \multicolumn{2}{c}{Intra-network}                      & \multicolumn{2}{c}{Inter-network} \\
\multicolumn{1}{c}{}                                                                               & \multicolumn{1}{l}{RN-18} & \multicolumn{1}{l}{ViT-Ti} & RN-18           & ViT-Ti          \\ \hline
PWCCA                                                                                              & 100\%                     & 100\%                      & 12.50\%         & 8.33\%          \\
OPD                                                                                                & 100\%                     & 100\%                      & 19.17\%         & 18.33\%         \\
LCKA                                                                                               & 100\%                     & 100\%                      & 96.11\%         & 34.81\%         \\
TLM                                                                                                & 63.75\%                   & 24.17\%                    & 35.28\%         & 10.00\%         \\
DM (struct.)                                                                                       & 100\%                     & 100\%                      & 92.50\%         & 25.19\%         \\
DM (func.)                                                                                         & 100\%                     & 100\%                      & 66.39\%         & 11.85\%         \\ \hline
\end{tabular}
\caption {Accuracies of identifying the same layer within a network and the architecturally corresponding layers between networks based on maximum similarity.}
\label{tab:matching_layers}
\end{table}

\paragraph{Failed inter-network sanity check.}
We can see that task loss matching performs very poorly at identifying the corresponding layers between different networks.
However, PWCCA and OPD also struggle with this, while LCKA performs very well, confirming the results in \cite{kornblith2019similarity}.
Interestingly, structural direct matching (see \cref{sec:twoface}) achieves comparable results to LCKA and its functional version outperforms task loss matching as well.
Also, all methods struggle on the ViT architecture, giving further evidence to the findings of \cite{raghu2021do}.

\paragraph{Failed intra-network sanity check.}
More alarmingly, task loss matching often fails to indicate that a layer is most similar to itself within a network.
This is quite problematic because passing this sanity check can be considered
a bare minimum requirement for any definition of similarity.

This failure is due to the training process that creates the stitching transformation,
because the training is always initialized with the help of direct matching that does not fail this test.
Let us now explore this problem by looking at the similarity values obtained with task loss matching.

\subsection{A Closer Look at Intra-Network Similarities}

\Cref{fig:self_stitch_sim_large} shows a \emph{similarity matrix} corresponding to our ResNet-18 experiments.
The matrix shows the functional similarity according to task loss matching between every layer pair,
with the source and the target layers indicated on the vertical and horizontal axes, respectively.

Clearly, task loss matching indicates high similarity values between layers that are very far apart, as was also shown in \cite{hernandez2022model}.
More strikingly, the similarity between very far layers can be just as high as the similarity between the corresponding layer (which, in the case of self-stitching, is the same layer).

This is unexpected, as neural networks are thought to develop layers with different functions: earlier layers tend to be generic, while the final layers conform to the task at hand \cite{Lenc2019a}.
In the case of classifiers, the final layers tend to cluster representations in the activation space \cite{Goldfeld2019a, papyan2020prevalence, balogh2023functional, yang2023neurons}.

\paragraph{A possible explanation.}
To understand why task loss matching indicates a high similarity between these functionally different representations, we hypothesize that directly optimizing the stitcher for functional performance causes the stitching layer to create \emph{out-of-distribution (OOD)} representations that fool the receiver to achieve good accuracy.
This would be somewhat surprising because of the relatively weak affine transformations that form the stitching layer.
To verify our hypothesis, we perform an OOD analysis of stitching.

\section{Out-of-Distribution Representations}

In order to study whether the internal representations are out-of-distribution (OOD),
first, we need to select a suitable method for OOD detection and adapt it to
our setting.

\paragraph{Energy-based OOD detection.}
For classification tasks, energy-based OOD detection \cite{secretenergy-iclr20, Liu2020d} has proven to be a successful approach.
This method assigns an energy score---namely the negative $\text{LogSumExp}(\cdot)$ of the logits of the classifier---to each input sample and then finds an energy threshold that best separates the energy scores of in-distribution (ID) and OOD samples.
In our experiments, we adopt the framework of \cite{Liu2020d} who pre-train a classifier on ID samples and then fine-tune it with an auxiliary OOD dataset to separate the energy scores of ID and OOD samples.

\paragraph{A dedicated OOD detector for each layer.}
We are interested in the distribution of the activations of a given layer, as opposed to that of the input samples.
To this end, we train a dedicated OOD detector network for every layer of interest, using the energy-based method of \cite{Liu2020d}.
The training dataset of this detector network for layer $i$ consists of the target activations $\{ g_{\leq i}(x_k) \}_{k=1}^K$, where $x_k \sim p(x), k=1, ..., K$.
In this dataset, the label of the activation $g_{\leq i}(x_k)$ is the same as that of $x_k$.
This dataset is then considered in-distribution.
Similarly, the auxiliary dataset can be generated using the target activations of an OOD dataset.

\subsection{Experimental Setup}

We perform our experiments using the CIFAR-10 dataset, where OOD detection is known to work well.
We use the 300K Random Images dataset \cite{hendrycks2019oe} as the auxiliary OOD dataset for fine-tuning.
As OOD detectors, we use ResNet-18 models for convolutional representations and ViT-Ti models without the initial embedding sections for vision transformer representations.
We trained OOD detectors on the activations of all stitched layers of every stitched network separately.
Additional details are discussed in the Supplementary.

\subsection{Results}

After training the OOD detectors, we can apply them to test if the distribution of the stitched activations $\{ [T_\theta \circ f_{\leq i}](x_k) \}_{k=1}^K$ can
be separated from the target activations, and thus detected as OOD.
We report the separability of the target and the stitched representations by the OOD detector with the help of the area under the receiver operating characteristic curve (AUROC) metric.
The OOD AUROC values for each layer pair are presented as a heatmap for easy comparison with the similar accuracy heatmaps (for example, \cref{fig:self_stitch_ood_large}).
\Cref{fig:stitch_tlm_dm} present the results for the experimental settings we evaluated.

\subsubsection{TLM prefers OOD representations.}
\begin{figure*}[h]
    \centering
    \begin{tabular}{ccccc}
	    & TLM Similarity & TLM OOD & DM Similarity & DM OOD \\
        \rotatebox[origin=l]{90}{\hspace{0mm} INTRA ResNet-18} &
		\includegraphics[width=.22\textwidth]{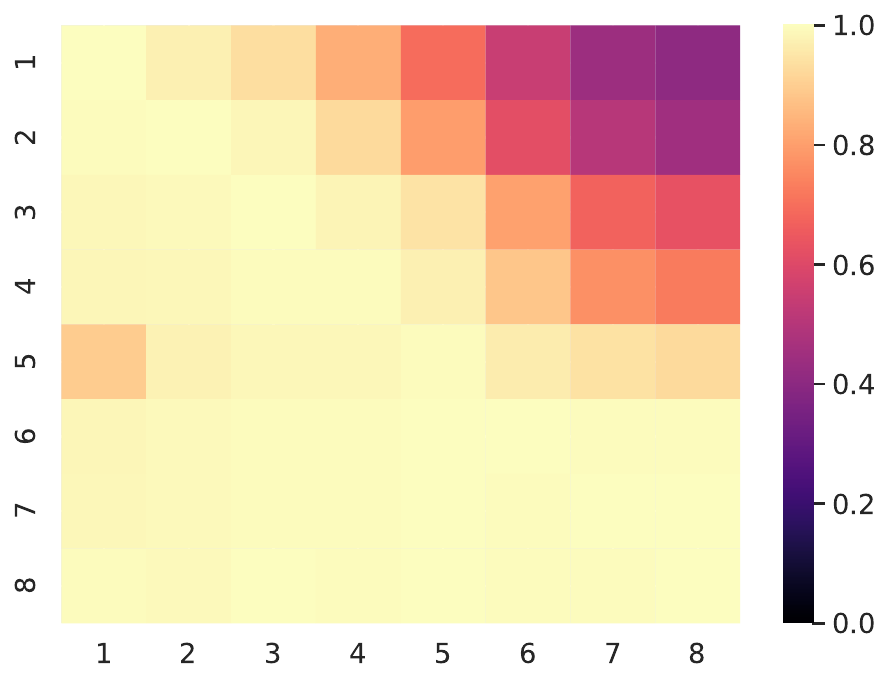} &
        \includegraphics[width=.22\textwidth]{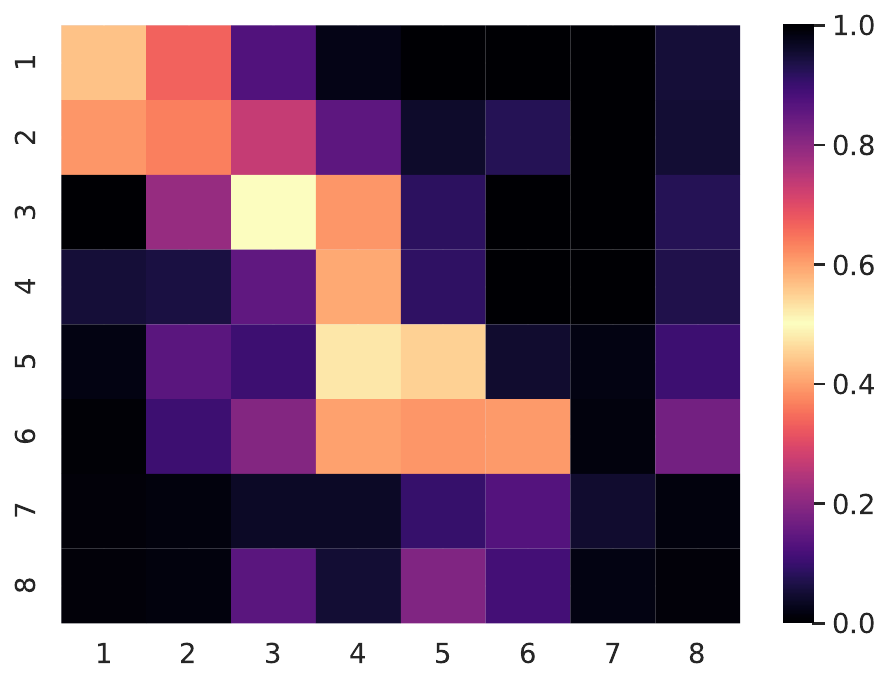} &
        \includegraphics[width=.22\textwidth]{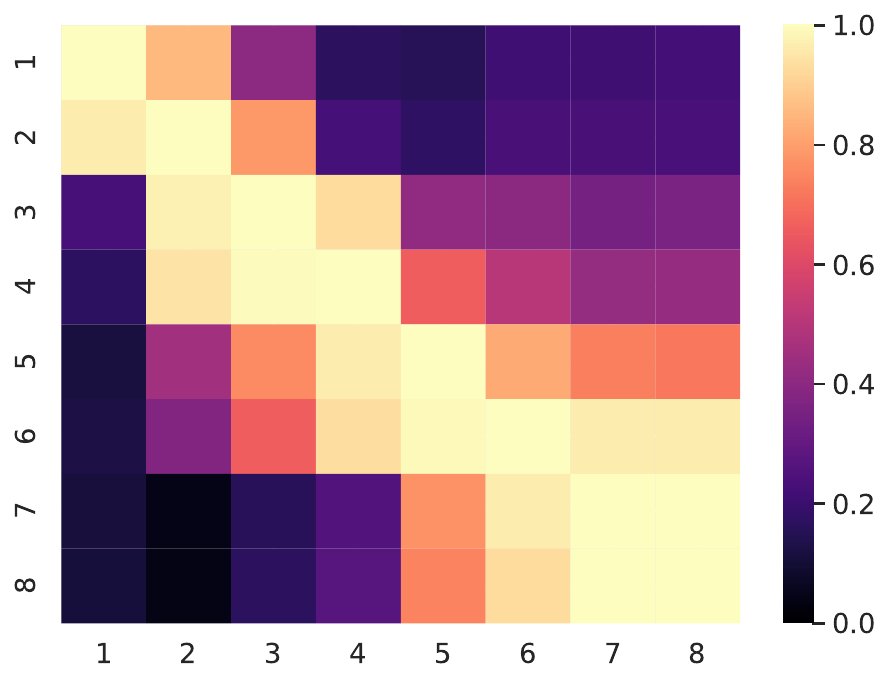} &
        \includegraphics[width=.22\textwidth]{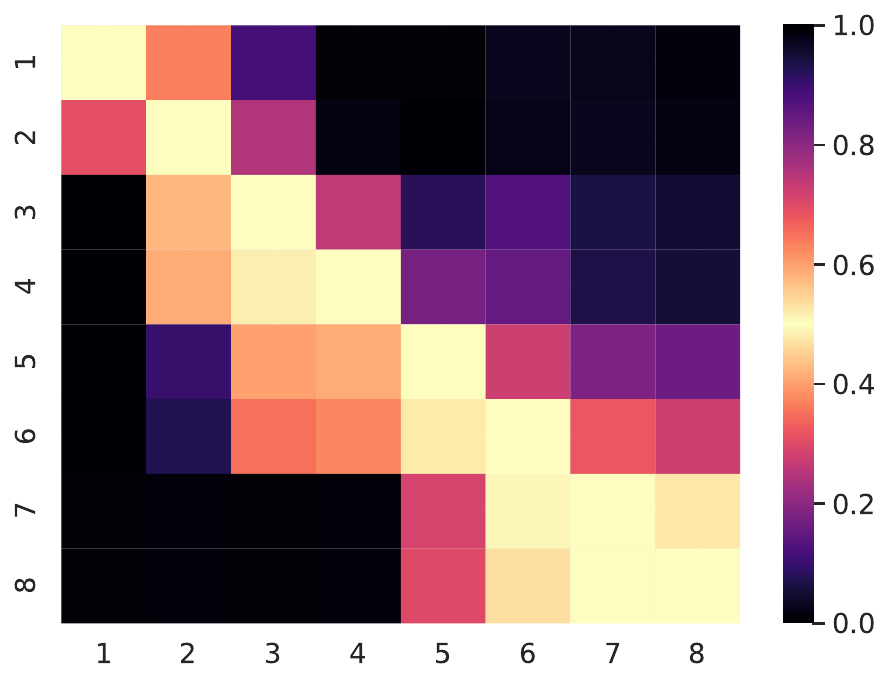} \\
        \rotatebox[origin=l]{90}{\hspace{3mm} INTRA ViT-Ti} &
        \includegraphics[width=.22\textwidth]{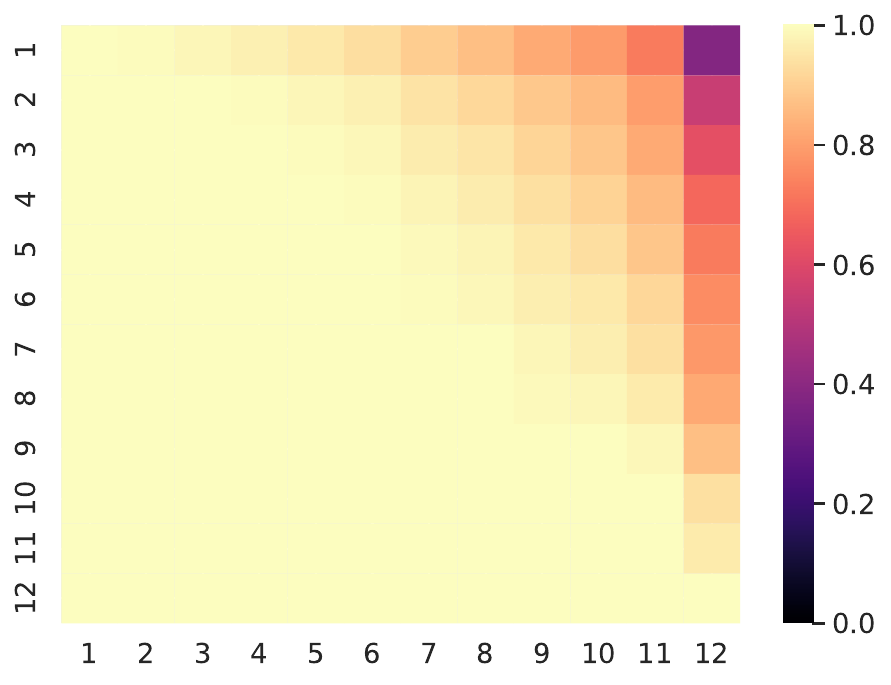} &
        \includegraphics[width=.22\textwidth]{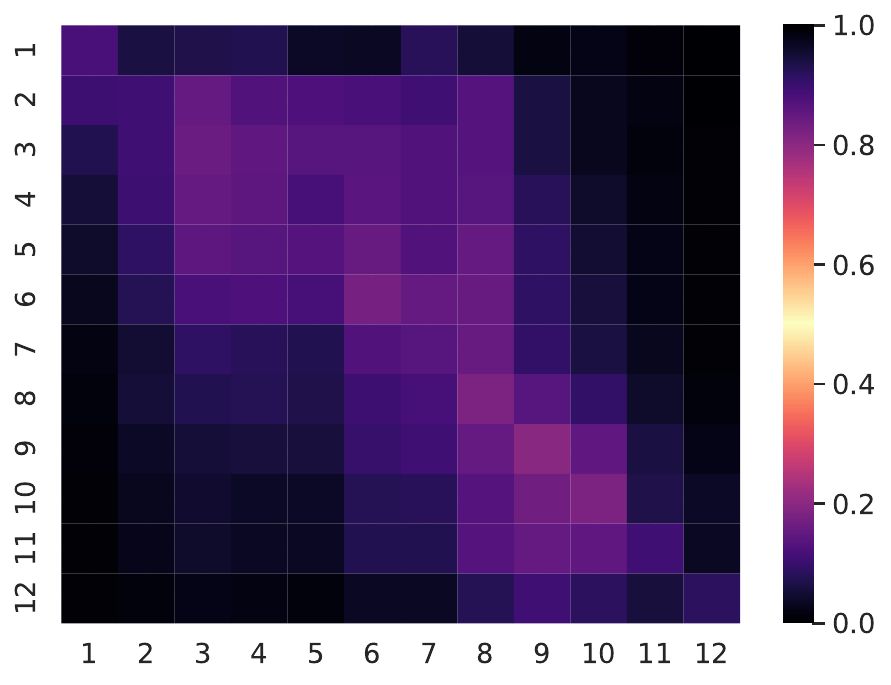} &
        \includegraphics[width=.22\textwidth]{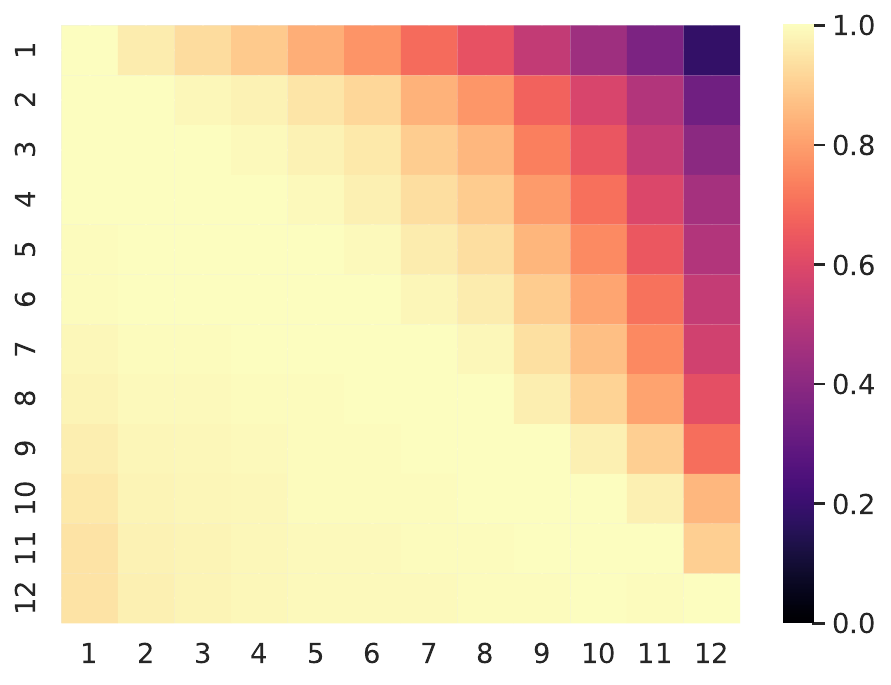} &
        \includegraphics[width=.22\textwidth]{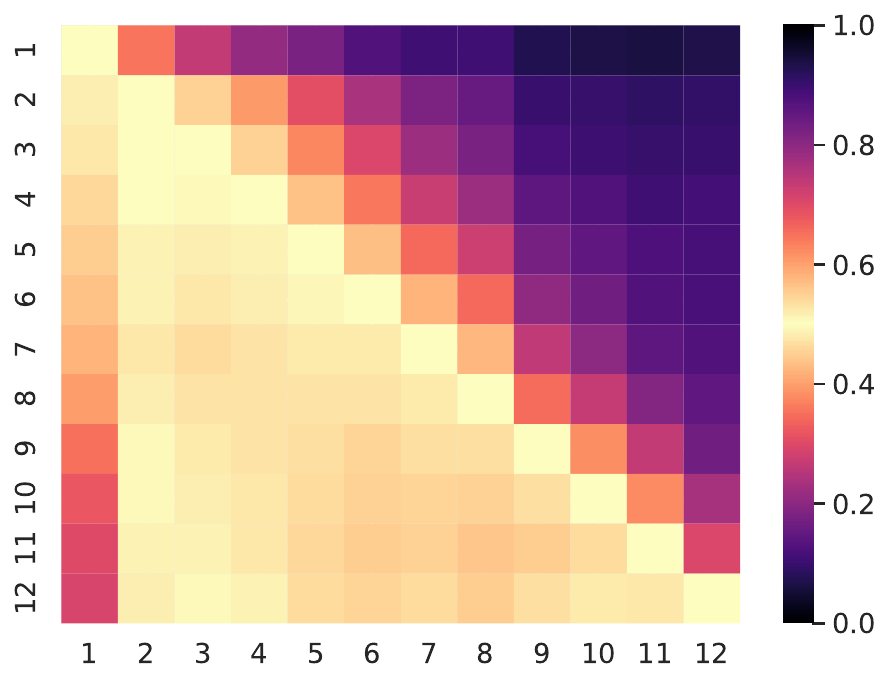} \\
        \rotatebox[origin=l]{90}{\hspace{0mm} INTER ResNet-18} &
        \includegraphics[width=.22\textwidth]{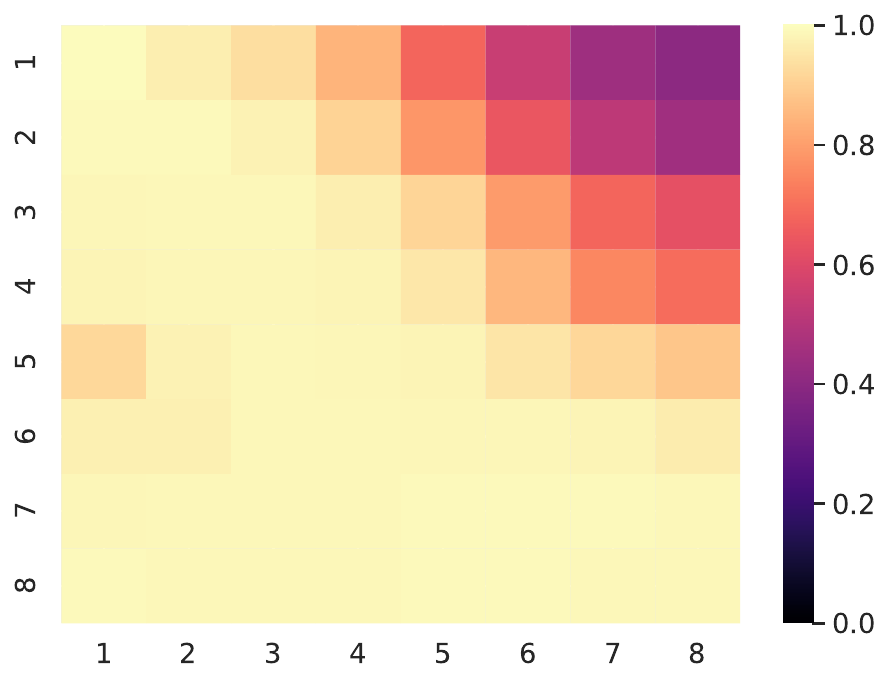} &
        \includegraphics[width=.22\textwidth]{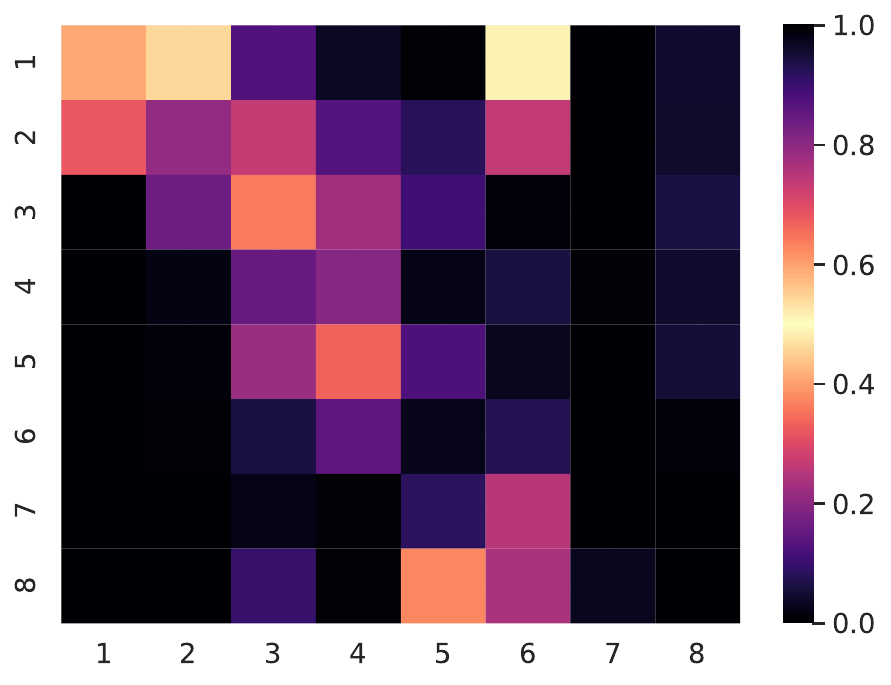} &
        \includegraphics[width=.22\textwidth]{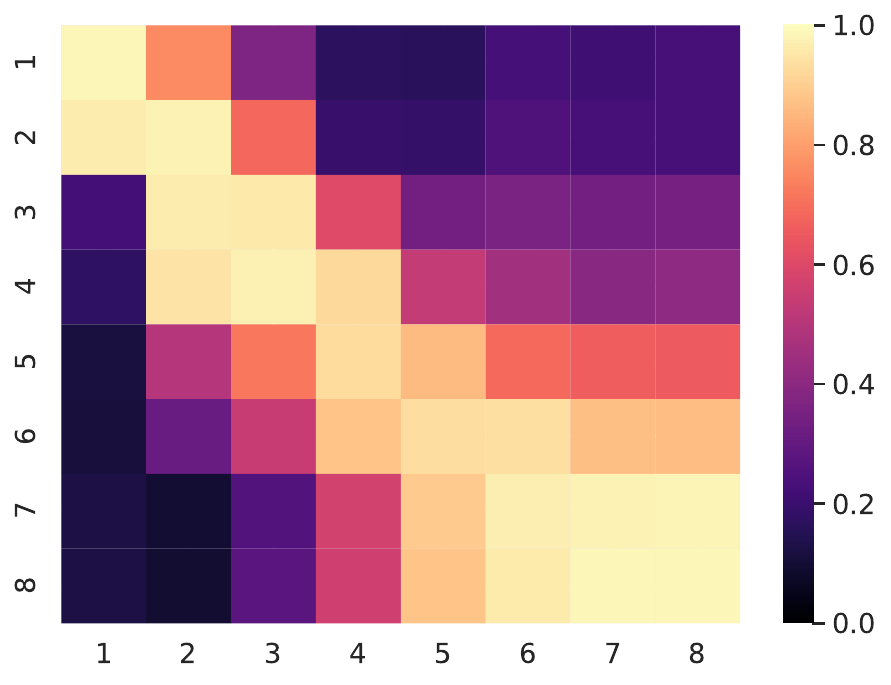} &
        \includegraphics[width=.22\textwidth]{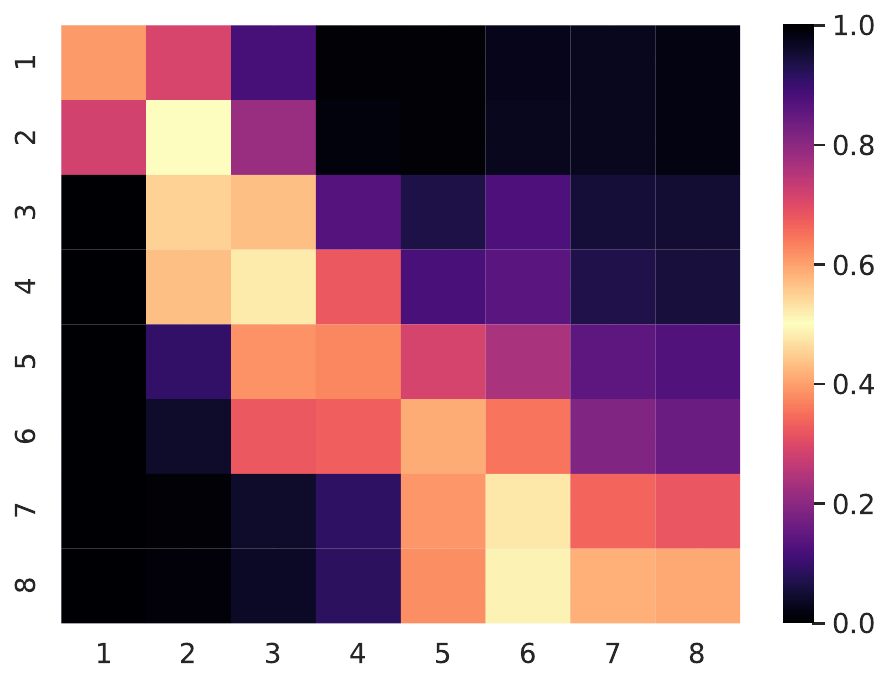} \\
        \rotatebox[origin=l]{90}{\hspace{3mm} INTER ViT-Ti} &
        \includegraphics[width=.22\textwidth]{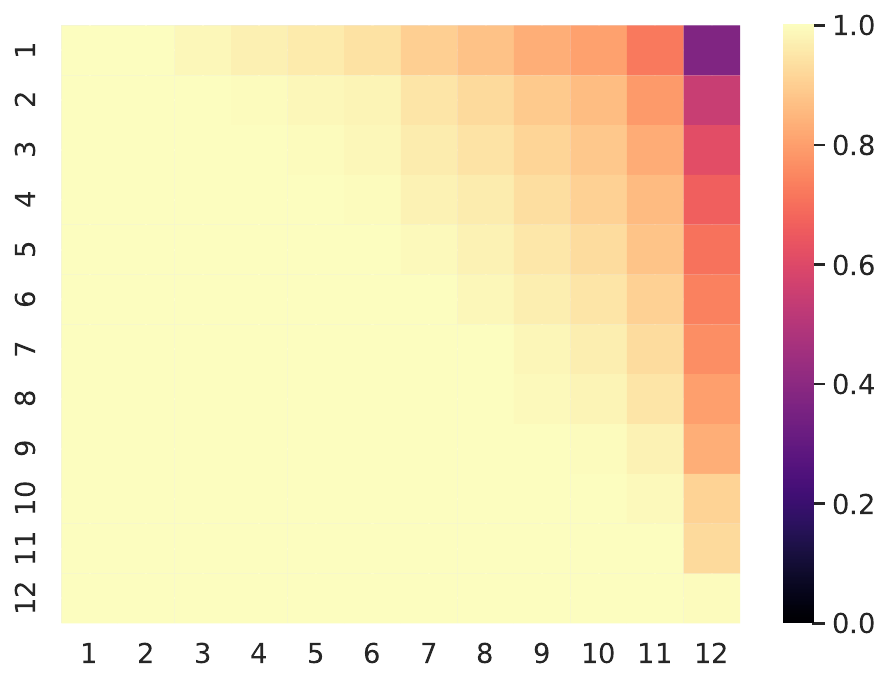} &
        \includegraphics[width=.22\textwidth]{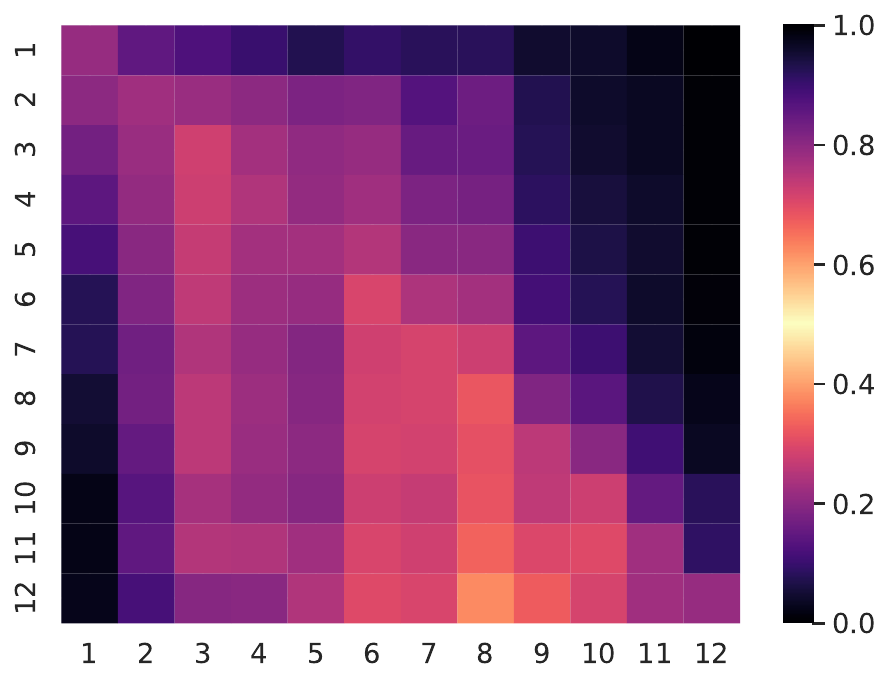} &
        \includegraphics[width=.22\textwidth]{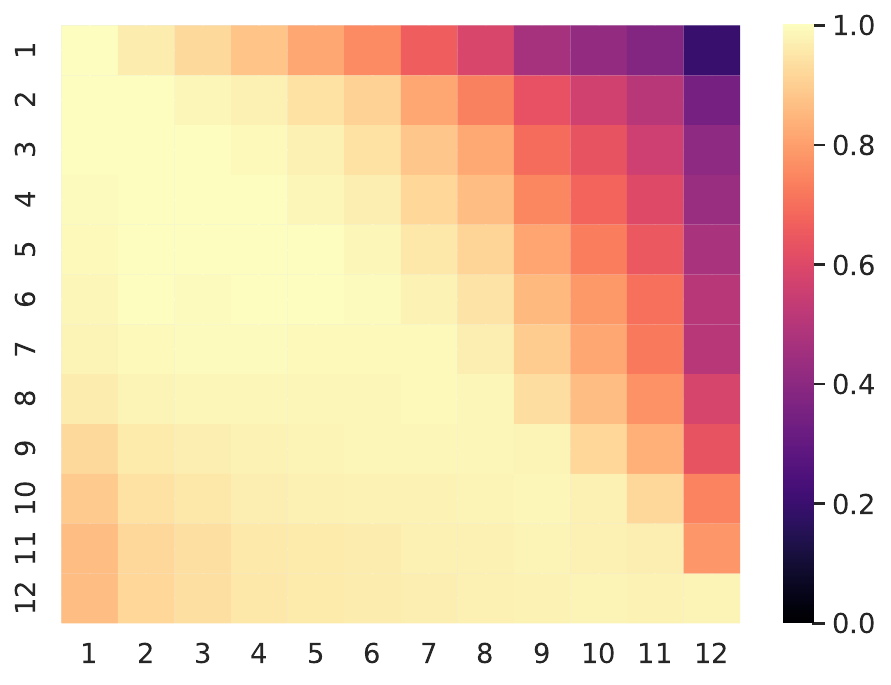} &
        \includegraphics[width=.22\textwidth]{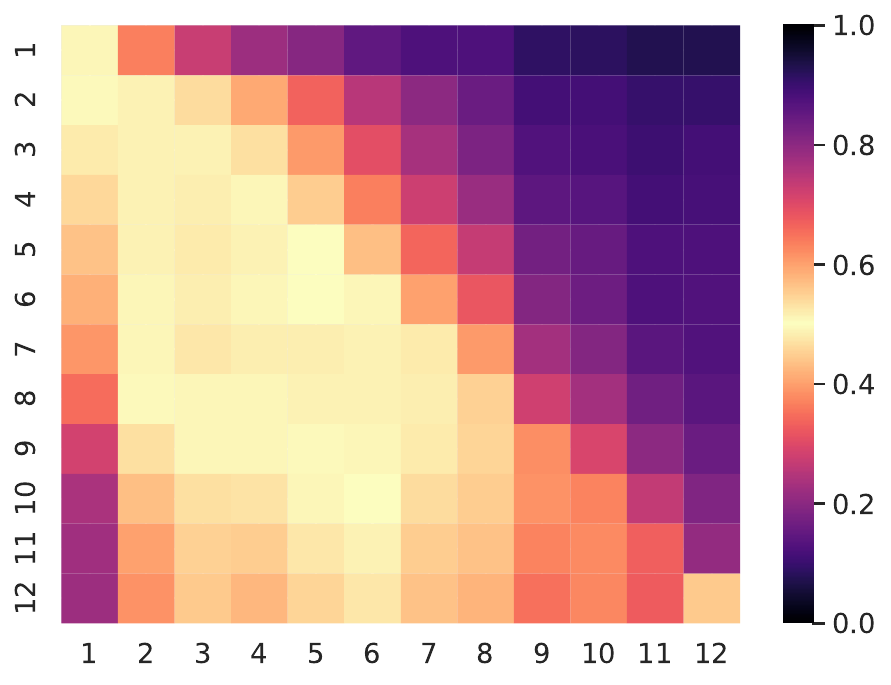} \\
    \end{tabular}
    \caption{
    Intra-network and inter-network stitching similarities (relative accuracy) and the corresponding OOD scores (AUROC)
    with task loss matching (TLM) and direct matching (DM).
	OOD scores close to 0.5 indicate in-distribution.
	The horizontal and vertical index is the target layer and the source layer, respectively.
    The stitched models were the same in each row.}
    \label{fig:stitch_tlm_dm}
\end{figure*}

The results confirm our earlier hypothesis, namely that TLM results in internal representations
that are predominantly OOD.
In fact, in most cases a complete separation is possible (see \cref{fig:self_stitch_energy_scores_large}, and the many
AUROC values close to 1 in \cref{fig:stitch_tlm_dm}).
The only exceptions are the cases where we match an early layer to itself (intra-network), or to the architecturally corresponding layer (inter-network).
At the same time, the relative accuracy of the stitched networks is predominantly excellent,
even when stitching distant layers.

What is even more striking is that TLM often creates \emph{OOD representations for architecturally matching layers} as well
for the ViT architecture and also for ResNets towards the end of the network, as shown in both \cref{fig:self_stitch_ood_large} and \cref{fig:stitch_tlm_dm}.
In the intra-network setting, even some identical layers are matched with OOD representations,
\emph{despite being initialized with the identity mapping}.

\paragraph{DM prefers ID representations.}
The objective of direct matching given in \cref{eq:matching-task} promotes creating in-distribution representations.
Indeed, in \cref{fig:stitch_tlm_dm} we can observe that, when direct matching indicates high functional similarity, it does so by creating ID representations.
Conversely, when the mapped representation is OOD, direct matching indicates low similarity.

\paragraph{OOD is undesirable.}
The results discussed here support the hypothesis that the reason behind TLM failing
the sanity checks in \cref{sec:sanity} is the emergence of OOD representations.
Since TLM can be considered a fine-tuning of DM transformations (since TLM learning is initialized with DM),
it is clear that TLM can modify the DM transformation significantly to gain maximum task
performance, leading to the violation of very natural sanity checks.
The OOD representations are therefore undesirable in a stitching-based functional similarity approach.
It should be noted, though, that TLM is still a suitable method
to provide evidence for the \emph{lack of functional similarity}.

\begin{table*}[]
\centering
\begin{tabular}{cccccccccccc}
\multirow{2}{*}{Task}     & \multirow{2}{*}{Model} & \multicolumn{2}{c}{LCKA} & \multicolumn{2}{c}{PWCCA}         & \multicolumn{2}{c}{OPD} & \multicolumn{2}{c}{DM (func.)}    & \multicolumn{2}{l}{DM (struct.)}  \\ \cline{3-12}
                            &                        & $\tau$        & $\rho$        & $\tau$             & $\rho$             & $\tau$            & $\rho$           & $\tau$             & $\rho$             & $\tau$             & $\rho$             \\ \hline
\multirow{2}{*}{CIFAR-10} & ResNet                 & 0.5687     & 0.7217     & 0.6276          & 0.7767          & 0.5484         & 0.7176        & \textbf{0.7896} & \textbf{0.9169} & 0.6583          & 0.8068          \\
                            & ViT                    & 0.4851     & 0.6320     & \textbf{0.8260} & \textbf{0.9453} & 0.7321         & 0.8621        & 0.7448          & 0.8885          & 0.7856          & 0.9057          \\ \hline
\multirow{2}{*}{SVHN}     & ResNet                 & 0.6924     & 0.8331     & 0.6913          & 0.8280          & 0.5740         & 0.7344        & \textbf{0.7617} & \textbf{0.8932} & 0.7051          & 0.8358          \\
                            & ViT                    & 0.7039     & 0.8379     & 0.6179          & 0.7823          & 0.6967         & 0.8460        & 0.6718          & 0.8636          & \textbf{0.7191} & \textbf{0.8789} \\ \hline
\multirow{2}{*}{ImageNet} & ResNet                 & 0.7762     & 0.8998     & \textbf{0.8315} & \textbf{0.9238} & 0.6696         & 0.8456        & 0.8096          & 0.9141          & 0.6631          & 0.7507          \\
                            & ViT                    & 0.5945     & 0.7518     & \textbf{0.8355} & \textbf{0.9458} & 0.6421         & 0.8146        & 0.7757          & 0.9209          & 0.7737          & 0.9025          \\ \hline
\end{tabular}
\caption{
    Kendall's $\tau$ and Spearman's $\rho$ rank correlations for the sensitivity test.
}
\label{tab:sensitivity}
\end{table*}

\begin{table*}[]
\centering
\begin{tabular}{cccccccccccc}
\multirow{2}{*}{Task}     & \multirow{2}{*}{Model} & \multicolumn{2}{c}{LCKA}  & \multicolumn{2}{c}{PWCCA} & \multicolumn{2}{c}{OPD} & \multicolumn{2}{c}{DM (func.)}    & \multicolumn{2}{c}{DM (struct.)}  \\ \cline{3-12}
    &                        & $\tau$             & $\rho$    & $\tau$         & $\rho$         & $\tau$        & $\rho$               & $\tau$             & $\rho$             & $\tau$             & $\rho$             \\ \hline
\multirow{2}{*}{CIFAR-10} & ResNet                 & 0.8590          & 0.9699 & 0.7285      & 0.8959      & 0.7678     & 0.9202            & 0.8571          & 0.9698          & \textbf{0.8760} & \textbf{0.9749} \\
    & ViT                    & 0.7189          & 0.8701 & 0.6502      & 0.7848      & 0.5314     & 0.6818            & \textbf{0.7818} & \textbf{0.9207} & 0.6639          & 0.8019          \\ \hline
\multirow{2}{*}{SVHN}     & ResNet                 & \textbf{0.9288} & 0.9916 & 0.7610      & 0.9136      & 0.9269     & \textbf{0.9916}   & 0.8895          & 0.9819          & 0.9006          & 0.9845          \\
    & ViT                    & 0.7148          & 0.8755 & 0.3015      & 0.4245      & $\approx 0$    & $\approx 0$           & \textbf{0.7415} & \textbf{0.8931} & 0.2898          & 0.4095          \\ \hline
\multirow{2}{*}{ImageNet} & ResNet                 & 0.8436          & 0.9516 & 0.6590      & 0.8214      & 0.5974     & 0.6784            & \textbf{0.8564} & \textbf{0.9643} & $\approx 0$         & $\approx 0$        \\
    & ViT                    & 0.7141          & 0.8645 & 0.4508      & 0.5965      & 0.4994     & 0.6356            & \textbf{0.7912} & \textbf{0.9179} & 0.4911          & 0.6443          \\ \hline
\end{tabular}
\caption{
    Kendall's $\tau$ and Spearman's $\rho$ rank correlations for the specificity test. The value $\approx 0$ indicates that the given correlation test is not statistically significant.
}
\label{tab:specificity}
\end{table*}

\section{Statistical Tests for Functional Similarity}

Here, we compare direct matching to structural similarity indices using the methodology of \cite{ding2021grounding}.
Ding et al.\ focus on the functional aspects of similarity.
Their main requirement is that \emph{a good notion of similarity should be correlated to functional similarity}.

They define a notion of functional similarity with the help of \emph{probing accuracy} \cite{probing}, where a simple
probing network is trained on a given representation.
If two representations achieve different accuracies, they are considered functionally different.
They then apply rank-correlation statistics to test whether a given similarity index is correlated with
functional similarity in a number of scenarios.

In the following, we describe the basic intuition behind the testing framework and the scenarios.
Further details are given in the Supplementary.

\paragraph{Sensitivity test.}
Here, the intuition is that low-rank approximations of a representation differ functionally
from the original representation (based on probing accuracy), so similarity indices should
also pick up on such differences.
That is, they should be \emph{sensitive to functional differences}.
This can be tested via generating a series of low-rank approximations of different ranks for a fixed layer,
and computing the correlation of functional (probing accuracy) and structural (similarity index) differences.

\paragraph{Specificity test.}
Here, like in our previous sanity checks, the intuition is that architecturally corresponding layers are functionally very close (based on
probing accuracy) so similarity indices should also indicate the highest similarity with the architecturally
corresponding layers, no matter which network they come from.
That is, they should be \emph{specific to functional differences}, and they should ignore non-functional differences such as random initialization.
This can be tested via generating several instances of the representations of every layer of a given architecture
(training with different random initializations) and again,
computing the correlations of functional and structural differences.

\paragraph{The testing framework.}
More formally, the testing framework of \cite{ding2021grounding} requires a set of representations $S$
and the probing accuracy $F$ defined over representations.
They select a reference representation
\begin{equation}
A = \argmax_{a\in S} F(a),
\end{equation}
and for every representation $B \in S$ they compute
$|F(A) - F(B)|$ and $d(A, B)$, where $d$ is a measure of dissimilarity (eg. OPD, 1-CKA, etc.).
Finally, they report the rank correlation between $|F(A) - F(B)|$ and $d(A, B)$, specifically, Kendall's $\tau$ and Spearman's $\rho$.

\subsection{Experimental Setup}

We present the key design choices here, the full set of details are discussed in the Supplementary.

\paragraph{Models.}
We study ResNet-18 and ViT-Ti models trained on the CIFAR-10, SVHN \cite{netzer2011reading} and ImageNet-1k \cite{ILSVRC15} datasets.
We trained 10 instances of both architectures on CIFAR-10 and SVHN and 5 instances of both architectures on ImageNet.
The training used the same hyperparameters within each scenario, but different initializations for the model instances.

\paragraph{Test settings.}
In the specificity test, we analyze all the 8 ResNet blocks and all the 12 ViT blocks.

In the sensitivity test, we include the last 4 ResNet blocks using low rank approximations of 13 and 14 different ranks for
blocks 5 and 6, and blocks 7 and 8, respectively.
Similarly, we studied the last 6 ViT blocks for CIFAR-10 and SVHN and the last 4 blocks for ImageNet at 16 different ranks.
The ranks ranged from full-rank to rank 1.

In the analysis of direct matching, we applied SVD-based low rank approximation (LRA)
to the source and target representations before computing the parameters of the optimal transformation.
We also used LRA on the source representations during evaluation.

Since our work is focused on image classifiers, we chose $F$ to be the linear probing accuracy of a representation \cite{alain2017understanding}.

\subsection{Results}

\Cref{tab:sensitivity} and \Cref{tab:specificity} show the results of the sensitivity and the specificity test, respectively.
We show the mean of the layer-wise results of the sensitivity test, as the results are consistent for individual layers.

\paragraph{Direct matching performs well.}
Functional direct matching achieves consistently high rank correlations with the functionality of interest in both tests.
In fact, it often has the highest rank correlations among all similarity measures, while being close to the leader in most of the other cases.
Therefore, we argue that direct matching can indeed serve as a meaningful measure of similarity, despite being a structural similarity at its core.
Based on direct matching's previously discussed tendency to create in-distribution representations, and its good performance in structural and functional benchmarks,
direct matching is a promising baseline for both structural and functional similarity.

\paragraph{The conclusions of Ding et al.\ revisited.}
As concluded by \cite{ding2021grounding}, LCKA performs poorly in the sensitivity test, while PWCCA achieves low rank correlations in the specificity test.
However, OPD---the similarity index recommended by Ding et al.---shows inconsistent results, which questions the conclusion that ODP would be a preferable baseline for functional evaluations.

\section{Conclusions and Limitations}

We argued that, although functional similarity is an important concept,
focusing purely on function can result in behavior that is strongly inconsistent with
basic sanity checks for similarity indices.
In fact, extending on the observation of \cite{hernandez2022model}, we even show that TLM-based model stitching (the method of
choice for functional similarity) can indicate that, for a given layer,
there are more similar layers than the layer itself, which is highly counter-intuitive.

We demonstrated that the reason behind these problems is that TLM-based stitching generates
highly OOD representations.
In fact, TLM can result in highly OOD representations even if architecturally identical layers are stitched.

Based on our empirical evaluation, we concluded that DM-based stitching combines structural and functional perspectives and
as such it generates in-distribution representations when matching is possible, while being able to assess functional
compatibility as well.

It has to be mentioned that our evaluation methodology is rather expensive.
\Cref{tab:matching_layers} alone requires computing 22,880 stitching transformations.
Altogether, we computed around 40,000 transformations, trained 60 models, 130 OOD detectors,
and 3590 probing layers.
Counting also our preliminary experiments, we used more than 1.5 GPU year's worth of computation on
a mixture of GPUs from RTX 2080TI to RTX A6000.

As for limitations, it would be useful to verify our claims on a wider set of models and applications,
including ones outside the image processing domain.
While in principle it is enough to demonstrate one problematic scenario to prove that a method
is not reliable, it is not clear how
generic the discovered problems are and how they depend on task-complexity and model complexity.
This, however, would require orders of magnitudes more resources than what is available to us.

\section*{Acknowledgements}
This work was supported by the University Research Grant Program of the
Ministry for Culture and Innovation from the source of the National Research, Development and Innovation Fund,
the European Union project RRF-2.3.1-21-2022-00004
within the framework of the Artificial Intelligence National Laboratory, and by
the project TKP2021-NVA-09, implemented with the support provided  by the Ministry of Culture
and Innovation of Hungary from the National Research, Development and Innovation Fund, financed under
the TKP2021-NVA funding scheme.

\bibliography{temp,raw}

\twocolumn[{
    \begin{center}
        \textbf{\Large Supplementary Material}\\[1cm]
    \end{center}
}]
\appendix

\section{Models}

In this section, we detail the architectures of the models we used in our experiments.
We trained models on the CIFAR-10 \cite{Krizhevsky2009a}, SVHN \cite{netzer2011reading} and ImageNet-1k \cite{ILSVRC15} classification tasks.

\paragraph{ResNets.}
In our experiments, we predominantly used the ResNet-18 \cite{deepresidual16} architecture.
For the CIFAR-10 and SVHN datasets, we used a ResNet-18 variant where the convolutional layer in the stem block is a $3 \times 3$ convolution with a stride of 1, and a 1-pixel zero-padding.
For the ImageNet dataset, we used the standard ResNet-18 variant where the same layer is a $7 \times 7$ convolution with a stride of 2, and a 3-pixel zero-padding.
The architecture was otherwise identical to the one provided in the torchvision library\footnote{https://github.com/pytorch/vision}.

\paragraph{Vision Transformers.}
We used the ViT-Ti \cite{Dosovitskiy2021a} architecture that has 12 transformer encoder blocks, 3 attention heads and an embedding dimension of 192.
For the CIFAR-10 and SVHN datasets, we trained transformers on the original $32 \times 32$ images, where the patch size was $4 \times 4$.
For the ImageNet dataset, we trained transformers on $256 \times 256$ images, where the patch size was $16 \times 16$.
As per standard practice, we used a dedicated \texttt{[cls]} token for classification. 
The architecture was identical to the one provided in the PyTorch Image Models library\footnote{https://github.com/huggingface/pytorch-image-models}.

\paragraph{OOD detectors.}
For OOD detection over internal activations, we used a modified ResNet-18 architecture for convolutional representations and a modified ViT-Ti architecture for transformer embeddings.
We modified the ResNet-18 architecture's stem block to start with a bilinear interpolation that resizes the input feature maps to the original spatial dimensions of the input ($32 \times 32$ in the case of CIFAR-10) and used the same $3 \times 3$ convolution in the stem block that was mentioned earlier.
We removed the initial embedding sections of the ViT architecture, which includes the patch embedding, position embedding, and appending a new \texttt{[cls]} token.
In other words, we only kept the transformer encoder blocks and the classifier head.
All other parts of the architectures were as we detailed previously.

We used full models as OOD detectors in all our experiments.
We found that partial models -- i.e. for the activations of $f_{\leq i}$, the architectural equivalent of $f_{>i}$ -- were good in classifying activations regardless of $i$, but were very poor at OOD detection for deeper layers.
We suspect that, for these deeper layers, the partial models simply did not have enough capacity to separate ID and OOD samples and thus used full models instead.

\paragraph{Linear probes.}
For probing convolutional representations, we applied average pooling on the input feature maps that reduced their spatial size to $1 \times 1$ and classified the flattened activations using a linear layer.
For transformer embeddings, we simply used a linear classifier on the \texttt{[cls]} token.
We included a bias term in all our linear classifiers.

\section{Training details}

In this section, we detail the training hyperparameters for all types of models we used in our experiments.

\subsection{Augmentations}
When training on the CIFAR-10 dataset, in all scenarios we used random horizontal flip and random crop as augmentations on the training images.
In certain settings, we relied on the CIFAR-5M dataset \cite{nakkiran2021the}, which is a synthetic dataset of 6 million CIFAR-10-like images.
We used the CIFAR-5M dataset by sampling 50K random images per epoch (equaling the size of the CIFAR-10 training set) without replacement and applying no augmentations.

When training on the SVHN dataset, we padded the images by 5 pixels using the edge pixels, applied random affine transformations and cropped the central $32 \times 32$ section of the resulting images.
The random affine transformations included rotations up to 5 degrees, scaling with a scaling factor between [0.9, 1.1], and vertical shearing by at most 5 degrees.

When training on the ImageNet-1k dataset, we used random resized cropping to $224 \times 224$ and random horizontal flipping on the training images, and bilinear resizing to $256 \times 256$ followed by cropping the central $224 \times 224$ section on the testing images.
We also normalized the training images, as per standard practice.

We also relied on 300K random images \cite{hendrycks2019oe} as the auxiliary dataset for training OOD detectors.
When using this dataset, we used the same augmentations as for CIFAR-10.

\subsection{Models}

\paragraph{CIFAR-10 ResNets.}
We trained ResNet-18 models on the CIFAR-10 dataset for 200 epochs with the Adam optimizer and with a batch size of 256.
The initial learning rate was $10^{-3}$, which was reduced to $10^{-4}$ after the $150^{\text{th}}$ epoch.
We also used weight decay with a coefficient of $10^{-5}$.
The average accuracy of the models we used in our benchmarks was 93.57\% with a standard deviation of 0.27\%.

\paragraph{CIFAR-10 ViTs.}
We trained ViT-Ti models using the CIFAR-5M dataset for 400 epochs, with the Adam optimizer, and the same batch size, initial learning rate and weight decay as with ResNets.
We reduced the initial learning rate to $10^{-4}$ after the $350^{\text{th}}$ epoch.
We also applied gradient clipping with an $\ell_2$-norm threshold of 1.
The average accuracy of the models we used in our benchmarks was 88.86\% with a standard deviation of 0.55\%.

\paragraph{SVHN ResNets.}
We used the exact same hyperparameters as we did in training ResNets on CIFAR-10.
The average accuracy of the models we used in our benchmarks was 95.94\% with a standard deviation of 0.14\%.

\paragraph{SVHN ViTs.}
The hyperparameters in this case are the same as for training ViTs on CIFAR-10, except we only trained ViTs on SVHN for 100 epochs and did not reduce the initial learning rate during training.
The average accuracy of the models we used in our benchmarks was 90.11\% with a standard deviation of 1.58\%.

\paragraph{ImageNet ResNets.}
We trained ResNet-18 models on the ImageNet-1k dataset for 90 epochs with a batch size of 1024.
We used the SGD optimizer with a momentum factor of 0.9, an initial learning rate of 0.1, which was divided by 10 after the $30^{\text{th}}$ and $60^{\text{th}}$ epochs, and used weight decay with a coefficient of $10^{-4}$.
The average accuracy of the models we used in our benchmarks was 69.30\% with a standard deviation of 0.57\%.

\paragraph{ImageNet ViTs.}
We trained ViT-Ti models for 100 epochs with a batch size of 1024.
We used the Adam optimizer with an initial learning rate of $10^{-3}$, which was reduced to $10^{-4}$ after the $75^{\text{th}}$ epoch, and used weight decay with a coefficient of $10^{-5}$.
Similar to earlier cases, we applied gradient clipping with an $\ell_2$-norm threshold of 1.
The average accuracy of the models we used in our benchmarks was 65.50\% with a standard deviation of 0.24\%.

\subsection{Task Loss Matching}

As mentioned before, we initialized all stitching layers with the direct matching method
\eachlabelcase{ {eq:matching-task} {given by \cref{eq:matching-task}} {given by eq.~(2) in the main text} }
with $K=100$ for every dataset.
We trained task loss matching on all datasets for 30 epochs with a batch size of 256.
In all scenarios, we used the Adam optimizer with a learning rate of $10^{-3}$ and a weight decay coefficient of $10^{-5}$.

\subsection{OOD Detectors}
As mentioned before, our goal is to train energy-based out-of-distribution (OOD) detectors on internal activations using the framework of \cite{Liu2020d}.
Here, we give a detailed description of our implementation.

Given a frozen network $g$ and one of its layers $g_i$, our goal is to train an OOD detector $f: \mathcal{A}_{g, i} \rightarrow \mathcal{Y}$.
The network $f$ should learn the same classification task as $g$, based on the representation of layer $i$ of $g$ as its input,
and it should be able to detect OOD inputs (thus, OOD representations in layer $i$) as well, using the method of \cite{Liu2020d}.

\paragraph{Pre-training.}
The first step is to pre-train $f$ to solve the classification problem $g$ was trained on.
This problem can be formalized as
\begin{equation}
\argmin_\theta \mathbb{E}_{(x_{\text{in}}, y_{\text{in}})}[\mathcal{L}(f_\theta(g_{\leq i}(x_{\text{in}})), y_{\text{in}})],
\end{equation}
where $\theta$ denotes the parameters of $f$, and $(x_{\text{in}}, y_{\text{in}})$ are random variables with the
training distribution (that is, the in-distribution).
The loss function $\mathcal L$ is the same loss used for training $g$.

In our experiments, we pre-trained the OOD detector for 100 epochs with a batch size of 256.
We used the Adam optimizer with an initial learning rate of $10^{-3}$ which was reduced to $10^{-4}$ after the $50^{\text{th}}$ epoch and a weight decay coefficient of $10^{-5}$.

\paragraph{Fine-tuning.}
The second step is to fine-tune $f$ to separate the energy scores of ID and OOD samples though solving the learning task
\begin{equation}
\argmin_\theta \mathbb{E}_{(x_{\text{in}}, y_{\text{in}}),x_{\text{out}}}[\mathcal{L}(f_\theta(g_{\leq i}(x_{\text{in}})), y_{\text{in}}) + \lambda \mathcal{L}_{\text{energy}}],
\end{equation}
where $\mathcal{L}_{\text{energy}}$ is given by
\begin{equation}
\begin{split}
\mathcal{L}_{\text{energy}} = & (\text{max}(0, \text{energy}_\theta(x_{\text{in}}) - m_{\text{in}})) ^ 2 + \\
                              & (\text{max}(0, m_{\text{out}} - \text{energy}_\theta(x_{\text{out}}))) ^ 2
\end{split}
\end{equation}
where $m_{\text{in}}$ and $m_{\text{out}}$ are hyperparameters that control the marginal energies of ID and OOD samples, respectively,
and the random variable $x_{\text{out}}$ follows some auxiliary OOD distribution.
In essence, $\mathcal{L}_{\text{energy}}$ penalizes ID samples that have higher energies than $m_{\text{in}}$ and OOD samples that have lower energies than $m_{\text{out}}$.
The function $\text{energy}_\theta(x)$ is given as
\begin{equation}
\text{energy}_\theta(x) = - \log \sum_j^C e^{f_{\text{logit}, j}(g_{\leq i}(x); \theta)},
\end{equation}
where $f_{\text{logit}, j}$ denotes the $j^{\text{th}}$ logit of $f$ and $C$ is the number of classes.
This energy function is also known as the negative $\text{LogSumExp}(\cdot)$ function.

In our experiments we used the CIFAR-5M dataset as the ID dataset and 300K random images as the OOD dataset in the fine-tuning step.
Using the OOD detector provided by \cite{Liu2020d} we confirmed that samples of the CIFAR-5M dataset are perfectly in-distribution when compared to samples from the CIFAR-10 dataset.

We fine-tuned our OOD detectors for 20 epochs with a batch size of 256 divided evenly between ID and OOD samples.
We used the Adam optimizer with an initial learning rate of $10^{-3}$ which was reduced to $10^{-4}$ after the $50^{\text{th}}$ epoch and a weight decay coefficient of $10^{-5}$.
We set the marginals as $m_{\text{in}} = -25$, $m_{\text{out}} = -7$ and the regularization weight $\lambda = 0.1$.

\subsection{Linear Probing}
All linear probes were trained for 10 epochs with the Adam optimizer, an initial learning rate of $10^{-3}$, which was reduced to $10^{-4}$ after the $5^{\text{th}}$ epoch and a weight decay coefficient of $10^{-5}$.
The batch size used in training was 256 for the CIFAR-10 and SVHN datasets and 1024 for the ImageNet dataset.

\section{Further Stitching Results}

\begin{figure*}[t]
\centering
\begin{tabular}{cccc}
\includegraphics[width=0.22\textwidth]{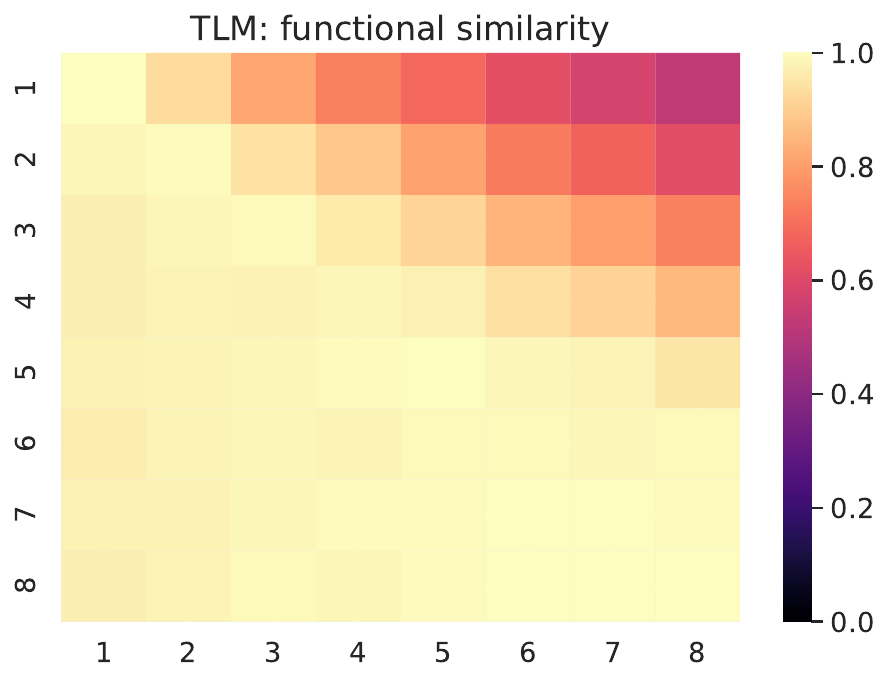} &
\includegraphics[width=0.22\textwidth]{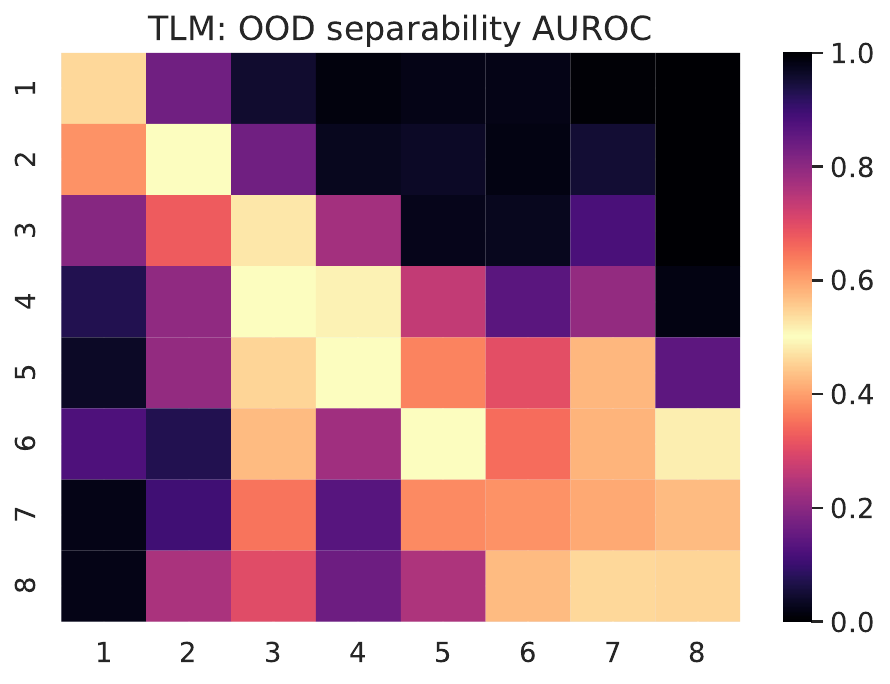} &
\includegraphics[width=0.22\textwidth]{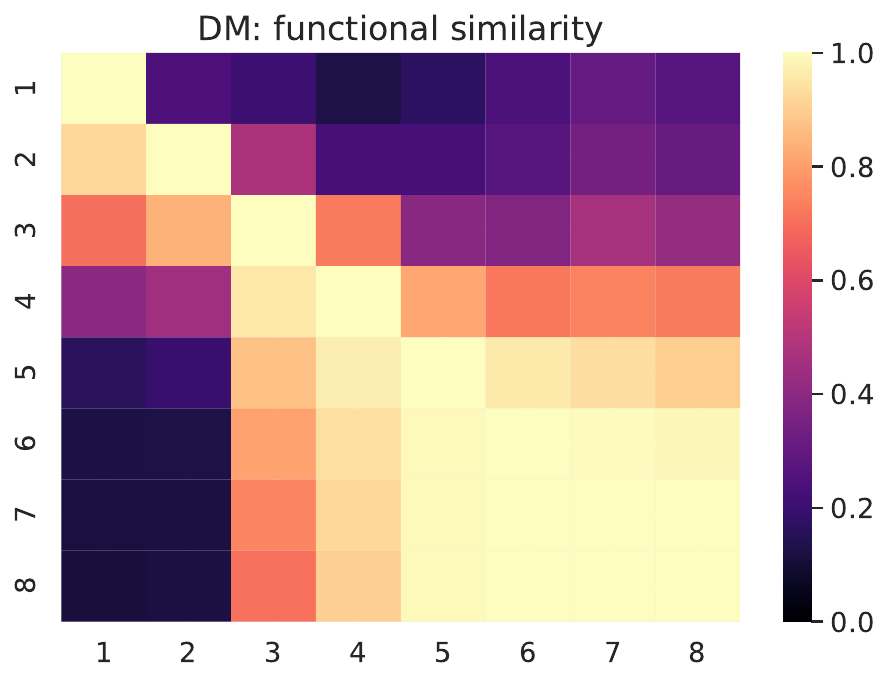} &
\includegraphics[width=0.22\textwidth]{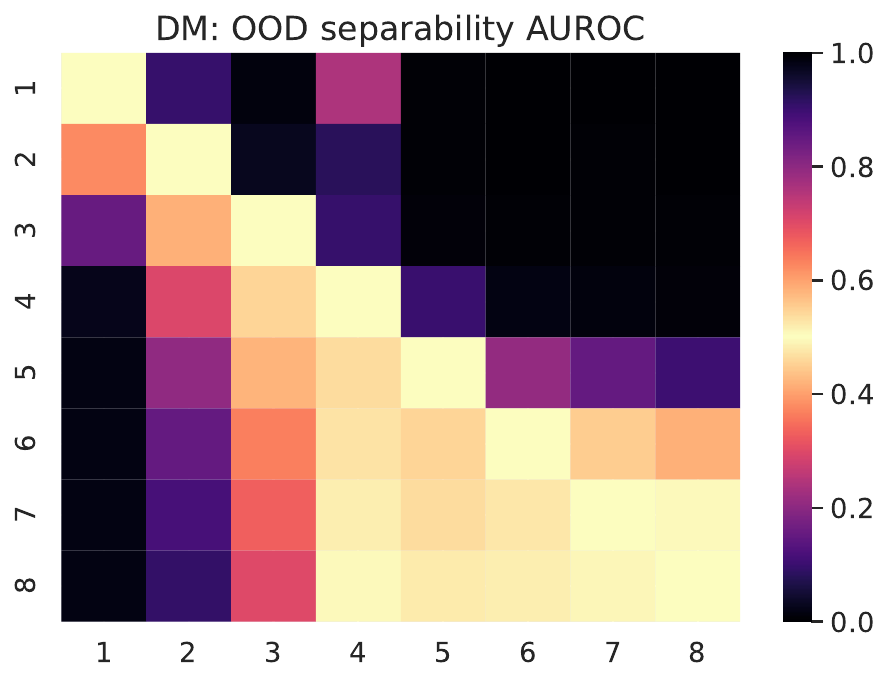} \\
\end{tabular}
\caption{
    Self-stitching a custom ResNet-18 with activations that have the same spatial dimensions.
    First two heatmaps: task loss matching similarity and its corresponding OOD AUROC heatmap.
    Second two heatmaps: direct matching similarity and its corresponding OOD AUROC heatmap.
}
\label{fig:samespatial}
\end{figure*}

\begin{figure}
\centering
\begin{tabular}{cc}
\includegraphics[width=0.21\textwidth]{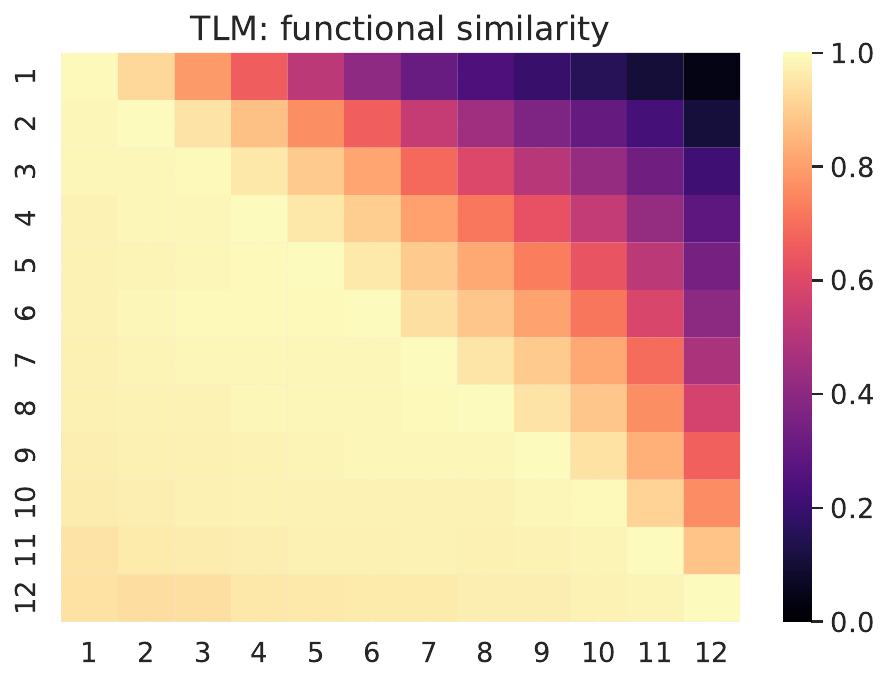} &
\includegraphics[width=0.21\textwidth]{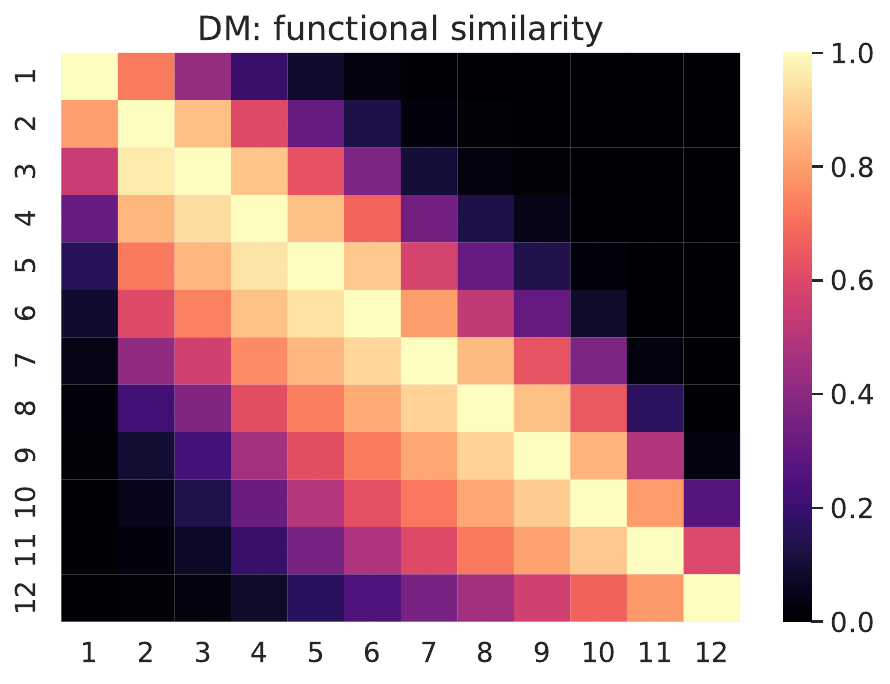} \\
\end{tabular}
\caption{Self-stitching a ViT-Ti model on the ImageNet dataset: task loss matching (left) and direct matching (right).}
\label{fig:stitch_imagenet}
\end{figure}

\paragraph{ImageNet results.}
To verify that task loss matching assigns high similarities to distant layers on more complex datasets as well,
we applied task loss matching to ViT models trained on ImageNet.
Due to the high computational requirement of task loss matching between every pair of layers over ImageNet, we only ran 5 epochs of training for every setting.
\Cref{fig:stitch_imagenet} shows the results of both task loss matching and direct matching.
It is clear that 5 epochs of training already results in high similarities between distant layers,
which is consistent with our results over the smaller datasets.

\paragraph{Difference in spatial dimension.}
Another hypothesis we considered is that the difference between the spatial dimensions of the stitched layers is responsible for the OOD behavior of task loss matching.
Although our results with ViTs, where all the the representations have the same size, do not support this hypothesis,
it is still possible that for convolutional architectures dimensionality has an influence.
For example, in the ResNet-18 experiments, when stitching to the first layer, there is a sharp change in OOD separability between using the first two or the remaining 6 layers
as a source layer, as shown in
\eachlabelcase{ {fig:self_stitch_energy_scores_large} {\cref{fig:self_stitch_energy_scores_large}} {fig. 1(c) in the main paper}}.
In this case, the representations of the first two layers have the same dimensionality but the remaining ones have different dimensionalities.

To test the effect of dimensionality, we created a custom ResNet architecture without pooling layers, where the spatial dimensions of all activations remained the same.
We also trained an OOD detector for every layer of this network.
\Cref{fig:samespatial} shows self-stitching similarities and their corresponding OOD separability heatmaps for this network.
It is clear that the change in spatial dimensions is not the reason behind the OOD phenomenon, because even with identical dimensions we can see that
distant layers can be stitched with task loss matching and this still results in OOD representations.
Direct matching achieves high accuracies with ID representations.

\section{Detailed Results of Statistical Tests}

\subsection{Structural Similarity Indices}
We used the LCKA implementation of \cite{kornblith2019similarity} and the PWCCA and OPD implementation of \cite{ding2021grounding}.

In order to get two-dimensional representations, we flattened the activations of each sample before computing LCKA.
For PWCCA and OPD, we took all spatial positions or tokens as samples.
In other words, in the latter case, a set of $n$ convolutional activations of size $c \times h \times w$ yielded a matrix of size $(n \cdot h \cdot w) \times c$ and a set of $n$ transformer embeddings of size $m \times d$ yielded a matrix of size $(n \cdot m) \times d$.
When dealing with convolutional representations with different spatial sizes, we applied adaptive average pooling to a spatial size of $1 \times 1$ as per the recommendation of \cite{raghu2017svcca}.

As for further preprocessing of the two-dimensional representations, we applied centering before computing LCKA and centering with normalization before computing PWCCA and OPD, just like \cite{ding2021grounding}.

\subsection{Sensitivity Test}

In the sensitivity test, we analyzed the last 4 blocks of ResNets and the last 6 blocks of ViTs, except for ImageNet, where we analyzed only the last 4 blocks.
We applied low-rank approximation to these layers' representations at the following ranks:
\begin{itemize}
    \item Blocks 5-6 of ResNets: \{ 256, 224, 192, 160, 128, 96, 64, 32, 16, 8, 4, 2, 1\}
    \item Blocks 7-8 of ResNets: \{ 512, 448, 384, 320, 256, 192, 128, 64, 32, 16, 8, 4, 2, 1\}
    \item ViT blocks: \{ 192, 176, 160, 144, 128, 112, 96, 80, 64, 48, 32, 16, 8, 4, 2, 1\}
\end{itemize}
Our choice of ranks shows a uniform decrease from full-rank at higher ranks (before rank 32) and a more fine-grained decrease towards rank 1, which conforms to the premise of the sensitivity test.

\Cref{tab:layerwise_sensitivity_rn18_cifar10,tab:layerwise_sensitivity_rn18_imagenet,tab:layerwise_sensitivity_rn18_svhn,tab:layerwise_sensitivity_vit_cifar10,tab:layerwise_sensitivity_vit_imagenet,tab:layerwise_sensitivity_vit_svhn}
show the layer-wise rank correlations in each setting with the associated $p$-values

\begin{table*}[t]
\centering
\begin{tabular}{cccccc}
\multicolumn{6}{c}{Kendall's $\tau$} \\ \hline
\hline
Layer & LCKA & PWCCA & OPD & DM (func.) & DM (struct.) \\ \hline
5&
0.6552 (2.17E-28)&
0.8446 (4.67E-46)&
0.5103 (7.45E-18)&
0.8336 (7.08E-45)&
0.8070 (3.72E-42)\\
6&
0.7064 (1.28E-32)&
0.5477 (1.33E-21)&
0.5556 (3.52E-22)&
0.8173 (6.01E-43)&
0.7134 (3.53E-33)\\
7&
0.4862 (2.38E-17)&
0.5591 (1.90E-22)&
0.5639 (8.35E-23)&
0.7799 (6.58E-42)&
0.5550 (4.26E-22)\\
8&
0.4271 (9.75E-14)&
0.5591 (1.90E-22)&
0.5639 (8.35E-23)&
0.7279 (8.40E-37)&
0.5576 (2.65E-22)\\ \hline
\multicolumn{6}{c}{\begin{tabular}{@{}c@{}}\\ Spearman's $\rho$\end{tabular}} \\ \hline
\hline
Layer & LCKA & PWCCA & OPD & DM (func.) & DM (struct.) \\ \hline
5 &
0.8286 (4.81E-34)&
0.9626 (1.83E-74)&
0.7187 (6.12E-22)&
0.9491 (4.21E-66)&
0.9323 (2.31E-58)\\
6&
0.8579 (7.75E-39)&
0.7111 (7.24E-23)&
0.7115 (6.75E-23)&
0.9349 (1.87E-59)&
0.8619 (1.42E-39)\\
7&
0.6212 (2.65E-16)&
0.7165 (2.47E-23)&
0.7202 (1.15E-23)&
0.9066 (1.45E-53)&
0.7148 (3.47E-23)\\
8&
0.5793 (6.49E-14)&
0.7165 (2.47E-23)&
0.7202 (1.15E-23)&
0.8767 (1.05E-45)&
0.7182 (1.74E-23)\\ \hline
\end{tabular}
\caption{
    Layer-wise Kendall's $\tau$ and Spearman's $\rho$ rank correlations (with $p$-values in parentheses) of the sensitivity test,
    performed with ResNet-18 models on the CIFAR-10 dataset.
}
\label{tab:layerwise_sensitivity_rn18_cifar10}
\end{table*}

\begin{table*}[t]
\centering
\begin{tabular}{cccccc}
\multicolumn{6}{c}{Kendall's $\tau$} \\ \hline
\hline
Layer & LCKA & PWCCA & OPD & DM (func.) & DM (struct.) \\ \hline
5 &
0.7189 (2.56E-31)&
0.6502 (6.34E-26)&
0.5314 (7.58E-18)&
0.7818 (9.96E-37)&
0.8490 (1.48E-46)\\
6&
0.8508 (9.83E-47)&
0.8632 (4.70E-48)&
0.7279 (1.12E-34)&
0.8668 (2.49E-48)&
0.8448 (4.54E-46)\\
7&
0.6509 (2.17E-29)&
0.7194 (7.45E-36)&
0.4892 (2.70E-17)&
0.6376 (3.54E-28)&
0.5447 (4.51E-21)\\
8&
0.5490 (1.68E-21)&
0.5323 (2.00E-20)&
0.5476 (2.14E-21)&
0.7608 (1.94E-39)&
0.5817 (6.27E-24)\\ \hline
\multicolumn{6}{c}{\begin{tabular}{@{}c@{}}\\ Spearman's $\rho$\end{tabular}} \\ \hline
\hline
Layer & LCKA & PWCCA & OPD & DM (func.) & DM (struct.) \\ \hline
5 &
0.8701 (4.58E-38)&
0.7848 (2.83E-26)&
0.6818 (1.01E-17)&
0.9207 (4.95E-50)&
0.9535 (1.56E-68)\\
6&
0.9637 (2.87E-75)&
0.9638 (2.23E-75)&
0.8809 (2.23E-75)&
0.9656 (8.97E-77)&
0.9448 (7.34E-64)\\
7&
0.7919 (2.31E-31)&
0.8667 (1.60E-43)&
0.6720 (9.92E-20)&
0.7910 (3.02E-31)&
0.7131 (4.93E-23)\\
8&
0.7065 (1.82E-22)&
0.6966 (1.21E-21)&
0.7028 (3.74E-22)&
0.8956 (2.20E-50)&
0.7321 (9.09E-25)\\ \hline
\end{tabular}
\caption{
    Layer-wise Kendall's $\tau$ and Spearman's $\rho$ rank correlations (with $p$-values in parentheses) of the sensitivity test,
    performed with ResNet-18 models on the SVHN dataset.
}
\label{tab:layerwise_sensitivity_rn18_svhn}
\end{table*}

\begin{table*}[t]
\centering
\begin{tabular}{cccccc}
\multicolumn{6}{c}{Kendall's $\tau$} \\ \hline
\hline
Layer & LCKA & PWCCA & OPD & DM (func.) & DM (struct.) \\ \hline
5 &
0.8288 (1.67E-22)&
0.9087 (1.02E-26)&
0.7606 (3.36E-19)&
0.8926 (1.16E-25)&
0.8981 (3.87E-26)\\
6&
0.8536 (9.27E-24)&
0.8805 (3.53E-25)&
0.6526 (1.56E-14)&
0.8340 (1.29E-22)&
0.8312 (1.33E-22)\\
7&
0.8004 (1.13E-22)&
0.8816 (3.70E-27)&
0.7002 (1.01E-17)&
0.9214 (8.38E-29)&
0.6883 (3.58E-17)\\
8&
0.6221 (2.64E-14)&
0.6552 (1.05E-15)&
0.5649 (4.68E-12)&
0.5906 (5.02E-13)&
0.2350 (4.04E-03)\\ \hline
\multicolumn{6}{c}{\begin{tabular}{@{}c@{}}\\ Spearman's $\rho$\end{tabular}} \\ \hline
\hline
Layer & LCKA & PWCCA & OPD & DM (func.) & DM (struct.) \\ \hline
5 &
0.9577 (8.45E-36)&
0.9837 (1.16E-48)&
0.9175 6.41E-27()&
0.9777 (2.16E-44)&
0.9804 (3.80E-46)\\
6&
0.9418 (1.61E-31)&
0.9613 (5.52E-37)&
0.8431 (1.24E-18)&
0.9369 (1.86E-30)&
0.9339 (7.76E-30)\\
7&
0.9229 (6.66E-30)&
0.9633 (1.46E-40)&
0.8634 (6.96E-22)&
0.9773 (1.56E-47)&
0.8281 (9.28E-19)\\
8&
0.7769 (2.66E-15)&
0.7869 (6.76E-16)&
0.7584 (2.89E-14)&
0.7646 (1.34E-14)&
0.2606 (2.94E-02)\\ \hline
\end{tabular}
\caption{
    Layer-wise Kendall's $\tau$ and Spearman's $\rho$ rank correlations (with $p$-values in parentheses) of the sensitivity test,
    performed with ResNet-18 models on the ImageNet dataset.
}
\label{tab:layerwise_sensitivity_rn18_imagenet}
\end{table*}

\begin{table*}[t]
\centering
\begin{tabular}{cccccc}
\multicolumn{6}{c}{Kendall's $\tau$} \\ \hline
\hline
Layer & LCKA & PWCCA & OPD & DM (func.) & DM (struct.) \\ \hline
7&
0.5712 (8.38E-27)&
0.8331 (4.45E-55)&
0.7107 (1.44E-40)&
0.7362 (2.15E-43)&
0.7720 (1.57E-47)\\
8&
0.4599 (1.63E-17)&
0.8397 (1.18E-85)&
0.7170 (4.91E-49)&
0.7385 (7.40E-54)&
0.7708 (5.49E-59)\\
9&
0.5548 (2.21E-25)&
0.8379 (1.05E-55)&
0.7757 (5.48E-48)&
0.8020 (3.68E-51)&
0.8117 (2.23E-52)\\
10&
0.5404 (3.74E-24)&
0.8488 (4.24E-57)&
0.7860 (3.19E-49)&
0.7228 (7.06E-42)&
0.8288 (1.69E-54)\\
11&
0.4945 (1.73E-20)&
0.8326 (5.22E-55)&
0.7576 (7.55E-46)&
0.7547 (2.21E-45)&
0.8236 (7.71E-54)\\
12&
0.2899 (5.54E-08)&
0.7638 (1.74E-46)&
0.6455 (1.06E-33)&
0.7144 (1.40E-40)&
0.7068 (4.66E-40)\\
\hline
\multicolumn{6}{c}{\begin{tabular}{@{}c@{}}\\ Spearman's $\rho$\end{tabular}} \\ \hline
\hline
Layer & LCKA & PWCCA & OPD & DM (func.) & DM (struct.) \\ \hline
7&
0.7201 (7.23E-27)&
0.9542 (1.11E-84)&
0.8518 (3.22E-46)&
0.8853 (2.09E-54)&
0.9046 (2.11E-60)\\
8&
0.6075 (1.63E-17)&
0.9555 (1.18E-85)&
0.8644 (4.91E-49)&
0.8833 (7.40E-54)&
0.9004 (5.49E-59)\\
9&
0.6898 (6.26E-24)&
0.9517 (6.65E-83)&
0.9052 (1.34E-60)&
0.9236 (1.10E-67)&
0.9309 (5.52E-71)\\
10&
0.6877 (9.76E-24)&
0.9569 (1.03E-86)&
0.9045 (2.33E-60)&
0.8741 (2.08E-51)&
0.9320 (1.54E-71)\\
11&
0.6493 (1.58E-20)&
0.9468 (1.00E-79)&
0.8763 (5.56E-52)&
0.8843 (3.81E-54)&
0.9270 (3.44E-69)\\
12&
0.4373 (7.38E-09)&
0.9065 (4.62E-61)&
0.7703 (1.09E-32)&
0.8806 (4.01E-53)&
0.8392 (1.18E-43)\\
\hline
\end{tabular}
\caption{
    Layer-wise Kendall's $\tau$ and Spearman's $\rho$ rank correlations (with $p$-values in parentheses) of the sensitivity test,
    performed with ViT-Ti models on the CIFAR-10 dataset.
}
\label{tab:layerwise_sensitivity_vit_cifar10}
\end{table*}

\begin{table*}[t]
\centering
\begin{tabular}{cccccc}
\multicolumn{6}{c}{Kendall's $\tau$} \\ \hline
\hline
Layer & LCKA & PWCCA & OPD & DM (func.) & DM (struct.) \\ \hline
7&
0.5401 (4.03E-24)&
0.6426 (1.83E-33)&
0.5555 (2.01E-25)&
0.6241 (1.20E-31)&
0.6156 (7.61E-31)\\
8&
0.6351 (1.01E-32)&
0.6444 (1.23E-33)&
0.6601 (3.25E-35)&
0.6280 (5.00E-32)&
0.6965 (5.27E-39)\\
9&
0.6851 (8.57E-38)&
0.6378 (5.62E-33)&
0.7204 (1.34E-41)&
0.6763 (7.43E-37)&
0.7239 (5.65E-42)\\
10&
0.7853 (4.41E-49)&
0.6145 (1.00E-30)&
0.7637 (1.61E-46)&
0.7274 (2.51E-42)&
0.7519 (4.11E-45)\\
11&
0.7721 (1.60E-47)&
0.5932 (9.44E-29)&
0.7312 (8.43E-43)&
0.6446 (1.27E-33)&
0.7605 (3.99E-46)\\
12&
0.8059 (1.31E-51)&
0.5749 (4.20E-27)&
0.7493 (7.55E-45)&
0.7305 (1.10E-42)&
0.7661 (8.49E-47)\\
\hline
\multicolumn{6}{c}{\begin{tabular}{@{}c@{}}\\ Spearman's $\rho$\end{tabular}} \\ \hline
\hline
Layer & LCKA & PWCCA & OPD & DM (func.) & DM (struct.) \\ \hline
7&
0.6992 (8.54E-25)&
0.8080 (3.87E-38)&
0.7245 (2.54E-27)&
0.8136 (4.83E-39)&
0.7925 (9.18E-36)\\
8&
0.7959 (2.96E-36)&
0.8122 (8.22E-39)&
0.8284 (1.27E-41)&
0.8451 (8.06E-45)&
0.8779 (2.19E-52)\\
9&
0.8171 (1.22E-39)&
0.8037 (1.91E-37)&
0.8724 (5.49E-51)&
0.8741 (2.10E-51)&
0.8881 (3.28E-55)\\
10&
0.9120 (5.01E-63)&
0.7774 (1.27E-33)&
0.9013 (2.71E-59)&
0.9077 (1.80E-61)&
0.9102 (2.35E-62)\\
11&
0.8923 (1.91E-56)&
0.7601 (2.19E-31)&
0.8688 (4.39E-50)&
0.8335 (1.45E-42)&
0.8974 (4.92E-58)\\
12&
0.9112 (9.54E-63)&
0.7323 (3.76E-28)&
0.8803 (4.86E-53)&
0.9075 (2.06E-61)&
0.9071 (2.92E-61)\\
\hline
\end{tabular}
\caption{
    Layer-wise Kendall's $\tau$ and Spearman's $\rho$ rank correlations (with $p$-values in parentheses) of the sensitivity test,
    performed with ViT-Ti models on the SVHN dataset.
}
\label{tab:layerwise_sensitivity_vit_svhn}
\end{table*}

\begin{table*}[t]
\centering
\begin{tabular}{cccccc}
\multicolumn{6}{c}{Kendall's $\tau$} \\ \hline
\hline
Layer & LCKA & PWCCA & OPD & DM (func.) & DM (struct.) \\ \hline
9&
0.5392 (1.45E-12)&
0.8449 (1.36E-28)&
0.6411 (3.85E-17)&
0.8135 (1.95E-26)&
0.7882 (4.40E-25)\\
10&
0.5880 ()&
0.8506 (1.17E-14)&
0.6652 (5.87E-29)&
0.8124 (1.98E-26)&
0.7941 (1.92E-25)\\
11&
0.6238 (2.64E-16)&
0.8429 (1.88E-28)&
0.6599 (4.61E-18)&
0.7794 (3.01E-24)&
0.7823 (9.98E-25)\\
12&
0.6268 (1.93E-16)&
0.8034 (5.37E-26)&
0.6021 (2.74E-15)&
0.6974 (7.13E-20)&
0.7302 (9.61E-22)\\
\hline
\multicolumn{6}{c}{\begin{tabular}{@{}c@{}}\\ Spearman's $\rho$\end{tabular}} \\ \hline
\hline
Layer & LCKA & PWCCA & OPD & DM (func.) & DM (struct.) \\ \hline
9&
0.6916 (1.23E-12)&
0.9586 (2.63E-44)&
0.8203 (1.28E-20)&
0.9541 (1.33E-42)&
0.9193 (2.46E-33)\\
10&
0.7470 (1.77E-15)&
0.9541 (1.30E-42)&
0.8356 (5.43E-22)&
0.9452 (1.18E-39)&
0.9163 (1.00E-32)\\
11&
0.7853 (6.49E-18)&
0.9474 (2.48E-40)&
0.8248 (5.28E-21)&
0.9216 (8.70E-34)&
0.9037 (1.81E-30)\\
12&
0.7834 (8.77E-18)&
0.9229 (4.49E-34)&
0.7777 (2.15E-17)&
0.8629 (7.89E-25)&
0.8706 (9.70E-26)\\
\hline
\end{tabular}
\caption{
    Layer-wise Kendall's $\tau$ and Spearman's $\rho$ rank correlations (with $p$-values in parentheses) of the sensitivity test,
    performed with ViT-Ti models on the ImageNet dataset.
}
\label{tab:layerwise_sensitivity_vit_imagenet}
\end{table*}

\subsection{Specificity Test}

\Cref{tab:specificity_rn18_cifar10,tab:specificity_rn18_imagenet,tab:specificity_rn18_svhn,tab:specificity_vit_cifar10,tab:specificity_vit_imagenet,tab:specificity_vit_svhn}
show the rank correlations for the specificity test in each setting with the associated $p$-values.
We note that on the ImageNet dataset, especially with the ViT-Ti architecture, intra-network similarities are very low towards the latter layers.
We hypothesize that this can be a result of task complexity significantly outweighing model complexity, resulting in brittle or less expressive representations that are therefore harder to match.
Another hypothesis is that, as a result of the low effective dimensionality of the representations towards the latter layers (again, compared to the complexity of the task), matching results in severely collapsed representations.
Despite this phenomenon, direct matching passes the functional test.
We argue that these seemingly contradicting results (namely, relatively low functional similarity, but still strongly consistent behavior)
are worth a further investigation from both theoretical and experimental angles.

\begin{table*}[t]
\centering
\begin{tabular}{lccccc}
& LCKA & PWCCA & OPD & DM (func.) & DM (struct.) \\ \hline
Kendall's $\tau$&
0.8590 (1.81E-29) &
0.7285 (1.18E-21) &
0.7678 (7.10E-24) &
0.8571 (2.50E-29) &
0.8760 (1.51E-30) \\
Spearman's $\rho$ &
0.9699 (1.31E-49) &
0.8959 (3.24E-29) &
0.9202 (1.67E-33) &
0.9698 (1.40E-49) &
0.9749 (1.23E-52) \\
\hline
\end{tabular}
\caption{
    Kendall's $\tau$ and Spearman's $\rho$ rank correlations (with $p$-values in parentheses) of the specificity test,
    performed with ResNet-18 models on the CIFAR-10 dataset.
}
\label{tab:specificity_rn18_cifar10}
\end{table*}

\begin{table*}[t]
\centering
\begin{tabular}{cccccc}
& LCKA & PWCCA & OPD & DM (func.) & DM (struct.) \\ \hline
Kendall's $\tau$&
0.7189 (2.56E-31) &
0.6502 (6.34E-26) &
0.5314 (7.58E-18) &
0.7818 (9.96E-37) &
0.6639 (6.11E-27) \\
Spearman's $\rho$&
0.8701 (4.58E-38) &
0.7848 (2.83E-26) &
0.6818 (1.01E-17) &
0.9207 (4.95E-50) &
0.8019 (3.63E-28) \\
\hline
\end{tabular}
\caption{
    Kendall's $\tau$ and Spearman's $\rho$ rank correlations (with $p$-values in parentheses) of the specificity test,
    performed with ViT-Ti models on the CIFAR-10 dataset.
}
\label{tab:specificity_vit_cifar10}
\end{table*}

\begin{table*}[t]
\centering
\begin{tabular}{lccccc}
& LCKA & PWCCA & OPD & DM (func.) & DM (struct.) \\ \hline
Kendall's $\tau$&
0.9288 (3.51E-34) &
0.7610 (1.72E-23) &
0.9269 (4.77E-34) &
0.8895 (1.88E-31) &
0.9006 (3.22E-32) \\
Spearman's $\rho$&
0.9916 (5.07E-71) &
0.9136 (3.23E-32) &
0.9916 (3.86E-71) &
0.9819 (3.56E-58) &
0.9845 (9.70E-61) \\
\hline
\end{tabular}
\caption{
 Kendall's $\tau$ and Spearman's $\rho$ rank correlations (with $p$-values in parentheses) of the specificity test,
 performed with ResNet-18 models on the SVHN dataset.
}
\label{tab:specificity_rn18_svhn}
\end{table*}

\begin{table*}[t]
\centering
\begin{tabular}{lccccc}
& LCKA & PWCCA & OPD & DM (func.) & DM (struct.) \\ \hline
Kendall's $\tau$&
0.7148 (5.67E-31) &
0.3015 (1.05E-06) &
-0.0507 (0.4116) &
0.7415 (3.40E-33) &
0.2898 (2.75E-06) \\
Spearman's $\rho$&
0.8755 (4.49E-39) &
0.4245 (1.35E-06) &
-0.0958 (0.2977) &
0.8931 (9.33E-43) &
0.4095 (3.41E-06) \\
\hline
\end{tabular}
\caption{
 Kendall's $\tau$ and Spearman's $\rho$ rank correlations (with $p$-values in parentheses) of the specificity test,
 performed with ViT-Ti models on the SVHN dataset.
}
\label{tab:specificity_vit_svhn}
\end{table*}

\begin{table*}[t]
\centering
\begin{tabular}{lccccc}
& LCKA & PWCCA & OPD & DM (func.) & DM (struct.) \\ \hline
Kendall's $\tau$&
0.8436 (1.77E-14) &
0.6590 (2.12E-09) &
0.5974 (5.65E-08) &
0.8564 (9.02E-15) &
-0.0603 (0.5839) \\
Spearman's $\rho$&
0.9516 (4.57E-21) &
0.8214 (8.38E-11) &
0.6784 (1.50E-06) &
0.9643 (1.58E-23) &
-0.1449 (0.3722) \\
\hline
\end{tabular}
\caption{
 Kendall's $\tau$ and Spearman's $\rho$ rank correlations (with $p$-values in parentheses) of the specificity test,
 performed with ResNet-18 models on the ImageNet dataset.
}
\label{tab:specificity_rn18_imagenet}
\end{table*}

\begin{table*}[t]
\centering
\begin{tabular}{lccccc}
& LCKA & PWCCA & OPD & DM (func.) & DM (struct.) \\ \hline
Kendall's $\tau$&
0.7141 (7.52E-16) &
0.4508 (3.59E-07) &
0.4994 (1.72E-08) &
0.7912 (4.77E-19) &
0.4911 (2.98E-08) \\
Spearman's $\rho$&
0.8645 (5.69E-19) &
0.5965 (4.92E-07) &
0.6356 (4.88E-08) &
0.9179 (5.97E-25) &
0.6443 (2.79E-08) \\
\hline
\end{tabular}
\caption{
 Kendall's $\tau$ and Spearman's $\rho$ rank correlations (with $p$-values in parentheses) of the specificity test,
 performed with ViT-Ti models on the ImageNet dataset.
}
\label{tab:specificity_vit_imagenet}
\end{table*}

\end{document}